\pdfoutput=1
\documentclass{article}

\PassOptionsToPackage{numbers, compress}{natbib}

\usepackage[main, final]{neurips_2026}
\usepackage{caption}
\usepackage{amsthm}

\usepackage[utf8]{inputenc} %
\usepackage[T1]{fontenc}    %
\usepackage{hyperref}       %
\usepackage{url}            %
\usepackage{booktabs}       %
\usepackage{amsfonts}       %
\usepackage{nicefrac}       %
\usepackage{microtype}      %
\usepackage{xcolor}         %

\usepackage{graphicx}
\usepackage{algpseudocode}
\usepackage{algorithm}
\usepackage{booktabs}
\usepackage{multirow}
\usepackage{amsmath}
\usepackage[capitalise]{cleveref}

\usepackage[english]{babel}

\newtheorem{proposition}{Proposition}

\title{Householder Motion Embedding: Noise Modulation for Motion-Controllable Video Generation}

\author{%
  Tao Jun Lin$^{1,2}$\thanks{Project Lead} \quad
  Jinhong Ni$^{1}$ \quad
  Ruyi Zha$^{1}$ \quad
  Yifu Wang$^{2}$ \\
  \bfseries Xibin Song$^{2}$ \quad
  Pan Ji$^{2}$ \quad
  Hongdong Li$^{1,3}$ \\[4pt]
  $^{1}$The Australian National University \quad
  $^{2}$Vertex Lab \\
  $^{3}$Amazon Inc.\\
  \texttt{\{taojun.lin, hongdong.li\}@anu.edu.au}
}

\begin{document}
{
\maketitle
\begin{center}
    \includegraphics[width=\linewidth]{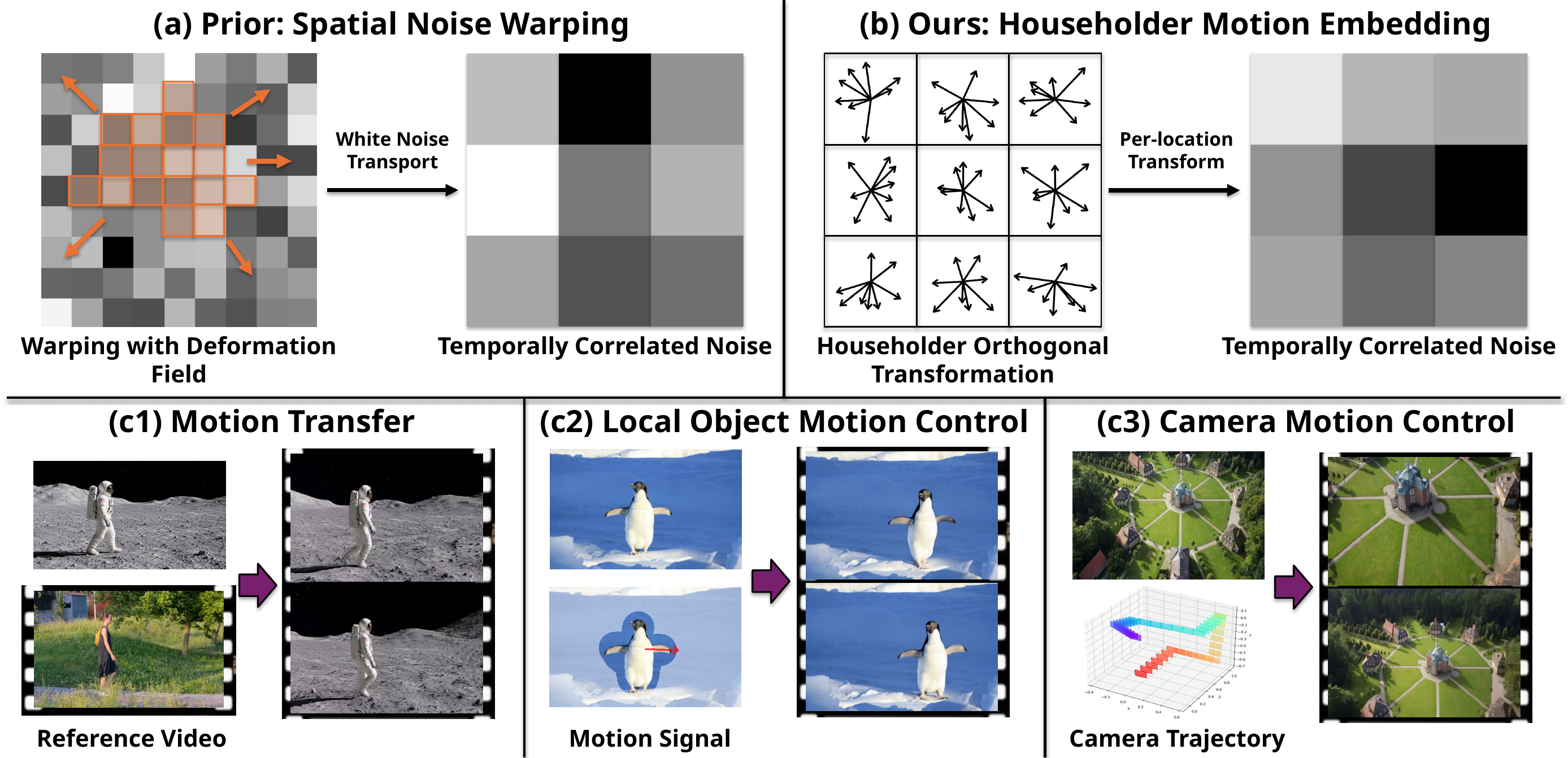}
\end{center}
\captionsetup{type=figure}
\captionof{figure}{We propose \textbf{Householder Motion Embedding (HME)}, a simple noise-modulation method for motion-controllable diffusion video generation. \textbf{(a)} Prior methods rely on \textbf{spatial noise warping} driven by motion fields. \textit{They preserve spatial Gaussianity only through costly correction steps.} \textbf{(b)} \textbf{HME} instead applies \textbf{per-location Householder transformations in channel space} to the noise. \textit{It preserves spatial Gaussianity by construction, with a $2.7\times$ speed-up and better motion adherence.} \textbf{HME} naturally supports \textbf{(c1)} motion transfer, \textbf{(c2)} local object motion control, and \textbf{(c3)} camera motion control. }
\label{fig:teaser}
}

\begin{abstract}

Diffusion models for video generation have achieved remarkable success, yet precise motion control under explicit user guidance remains challenging. Recent noise-warping methods introduce temporal correlations by warping the noise according to the user-provided motion signal.  However, standard spatial warping generally disrupts the noise’s i.i.d. Gaussian structure, requiring distribution correction, such as Jacobian-based correction, but at a substantial computational overhead.  We propose \textbf{Householder Motion Embedding (HME)}, which encodes the motion signal into the noise through a sequence of Householder reflections that together form an orthogonal transformation applied to each location’s channel vector. \textbf{HME} exactly preserves the spatial Gaussian structure within each frame while introducing cross-frame correlations that promote temporally consistent video. \textbf{HME} is agnostic to the network architecture and integrates directly into standard video diffusion fine-tuning pipelines. Experiments on different scenarios show that \textbf{HME} matches or outperforms the state of the art, while achieving a 2.7× speedup over noise warping. These results demonstrate that per-location Householder transformations in channel space offer a principled and efficient alternative to noise warping for motion-controllable video generation. 
\end{abstract}

\section{Introduction}
\label{sec:intro}
Diffusion based models have recently achieved remarkable success in video generation, demonstrating strong visual fidelity and semantic coherence across temporal dimension~\cite{yang2024cogvideox, kong2024hunyuanvideo, videoworldsimulators2024, blattmann2023stable, guo2024animatediff, zheng2024open, wan2025wan, gao2025seedance}. However, generating temporally consistent videos under explicit motion controls (modeling complex camera motion or scene dynamics) remains a fundamental challenge. Existing approaches~\cite{zhang2024controlvideo, shi2024motioni2v, jain2024peekaboo, mou2024revideo} often suffer from temporal drift, geometric inconsistency, or limited controllability when motion deviates from 
patterns observed during training.

A common wisdom to inject motion cues into the neural network architecture is through feature modulation~\cite{zhang2025motionpro,resslerdismo} or motion conditioning tokens~\cite{geng2025motion, wang2024motionctrl, he2024cameractrl, wu2024draganything}, which operate downstream of the stochastic generation process and leave the noise motion-agnostic.  An alternative is to modify the noise directly.  Building on the observation by PYoCo~\cite{ge2023pyoco} that shared noise components across frames improve temporal coherence, HIWYN~\cite{chang2024warped}, InfRes~\cite{deng2025infiniteresolution}, and GWTF~\cite{burgert2025go} demonstrate that constructing temporally correlated noise by transporting the a priori latent noise via diffeomorphic deformation fields parameterized by optical flow significantly enhances frame-to-frame consistency without modifying model architecture or training objectives. However, spatial resampling during this noise transport inherently destroys the i.i.d.\ Gaussian structure of the noise, since interpolation induces correlations across neighboring locations and alters the marginal distribution. This matters because the diffusion model is trained to denoise i.i.d.\ standard Gaussian noise: noise with spatial autocorrelation or altered variance falls outside this training distribution, which prior work has shown degrades generation quality~\cite{chang2024warped, deng2025infiniteresolution, burgert2025go}. Consequently, all existing noise-warping methods must introduce additional processing steps to restore spatial Gaussianity at the cost of non-trivial computational overhead.

In this paper, we propose \textbf{Householder Motion Embedding (HME)}, a noise-modulation framework that resolves this spatial-Gaussianity bottleneck without sacrificing motion fidelity. Rather than warping noise via spatial deformation fields, we modulate the noise using the intended motion field by per-location orthogonal transformations in channel space, parameterized as the product of Householder reflections~\cite{golub2013matrix}. The Householder transformation, composed of a small number of reflections, yields expressive orthogonal operators at low computational cost, while the orthogonality ensures distribution preservation by construction. 

Crucially, this distribution preservation guarantee is precisely what spatial noise warping methods lack. Modulating the noise through per-location Householder products therefore eliminates post-hoc correction entirely, requiring no Jacobian adjustments or distribution restoration steps, while inducing structured cross-frame temporal correlations that encode the geometric consistency of the underlying motion. Compared to deformation field based solutions, our method achieves exact spatial Gaussianity by construction, a $2.7\times$ speedup in noise processing time, and improved motion adherence on all three benchmarks.

Similar to existing noise warping methods, \textbf{HME} operates entirely at the noise initialization stage and requires no modifications to the denoising architecture, loss function, or training objective. The motion-embedded noise replaces the standard Gaussian noise at the start of the diffusion process, making the method plug-and-play across UNet-based and transformer-based video diffusion frameworks; we demonstrate it on the transformer-based CogVideoX. Our framework accepts any dense optical flow field as conditioning input regardless of its source: whether extracted from a reference video for motion transfer, rendered from a target camera 
trajectory for camera-controlled generation, or derived from user-specified drag interactions on the initial frame, thereby unifying diverse motion modalities within a single framework without motion-type-specific architectural modifications.

Contributions of this paper are as follows:
\vspace{-2mm}
\begin{enumerate}
    \item We propose a novel noise-modulation method for motion-controllable video diffusion that replaces spatial noise transport with per-location orthogonal transformations in latent channel space, parameterized as products of Householder reflections. Unlike existing noise-warping methods, this construction requires no post-hoc Gaussianity correction and operates in a single forward pass with no learnable parameters, making it a drop-in replacement for the noise initialization step in any video diffusion fine-tuning pipeline.
    \item We derive the cross-frame covariance of the proposed noise, showing that the dependence between frames is an orthogonal transformation determined by the local motion, and analyse how the number of reflections controls the expressivity of this encoding.
    \item We validate through extensive experiments on spatial Gaussianity preservation, motion-transfer and camera-controlled video generation benchmarks that \textbf{HME} matches or exceeds the state-of-the-art noise-warping baseline on nearly all motion-adherence and generation-quality metrics, while achieving a $2.7\times$ speedup in noise processing time.
\end{enumerate}

\vspace{-0.1in}
\section{Related Work}
\label{sec:related}
\vspace{-0.1in}

{\textbf{Motion-controllable video generation.}} Motion-controllable video generation methods broadly fall into three categories. Local object motion control methods~\cite{geng2025motion, jain2024peekaboo, yatim2024space} condition generation on user-specified trajectories via feature modulation or latent optimization, including Gaussian-filtered trajectory approaches~\cite{yin2023dragnvwa, wu2024draganything}, two-stage sparse-to-dense motion field estimation~\cite{shi2024motioni2v, niu2024mofa}, region-wise trajectory and mask conditioning~\cite{zhang2025motionpro}, and training-free latent optimization~\cite{namekata2025sgi2v}. Global camera control methods~\cite{kuang2024collaborative, xing2024dynamicrafter, wang2025bullettime} embed camera pose sequences into the denoising network through dedicated conditioning modules~\cite{he2024cameractrl, bahmani2025vd3d, wang2024motionctrl, yang2024direct, xu2024camco, zhang2024recapture} or learning Video-to-Video generation directly from imperfect 3D/4D renders~\cite{ren2025gen3c,gu2025diffusion2}, which require large posed-video datasets for training. Motion transfer methods~\cite{aira2024motioncraft,geyer2024tokenflow, ku2024anyv2v, gokmen2025ropecraft,pondaven2025ditflow} propagate motion signals from a reference clip without additional supervision, either through temporal attention cloning~\cite{ling2025motionclone}, zero-shot feature guidance~\cite{yatim2024space}, inference-time schedule adaptation~\cite{singer2025time}, motion disentanglement~\cite{resslerdismo}, or trajectory and optical flow conditioning~\cite{geng2025motion, li2025imageconductor, wang2024videocomposer}. Across all three paradigms, motion signals are injected downstream of the stochastic initialization, leaving the noise distribution motion-agnostic. Our method addresses this by modulating the noise through per-location orthogonal transformations, which preserve the i.i.d.\ Gaussian structure by construction while inducing cross-frame correlations that align per-frame noise distributions along the motion trajectory.

{\textbf{Noise Warping for Temporally Consistent Video Generation.}}  Regardless of architectural difference between UNet-based and DiT-based video diffusion models, both are initialized from Gaussian noise whose correlation across frames can be influential to the temporal consistency of the generated output. PYoCo~\cite{ge2023pyoco} first demonstrates that correlating the noise initialization across frames through a mixed noise prior substantially improves temporal coherence, motivating subsequent work on flow-guided noise warping. 
 HIWYN~\cite{chang2024warped} introduces integral noise, reinterpreting warped noise samples as integrals of a continuous underlying noise field to preserve spatial Gaussianity during deformation field guided transport, at the cost of higher computational complexity. InfRes~\cite{deng2025infiniteresolution} resolves this by deriving an equivalent infinite-resolution formulation through Brownian bridge increments, reducing overhead by orders of magnitude while matching HIWYN's theoretical guarantees. GWTF~\cite{burgert2025go} further improves efficiency with a linear-time warping algorithm and demonstrates motion-controllable fine-tuning of modern video diffusion models across object motion, camera control, and motion transfer. Despite these advances, all existing methods share a fundamental limitation: spatial resampling during deformation-field-based transport disrupts the i.i.d.\ Gaussian noise structure, necessitating costly post-hoc correction at higher resolutions. Our method resolves this bottleneck by replacing spatial transport with per-location orthogonal transformations in channel space that preserve spatial Gaussianity exactly by construction.
 
Our method makes use of Householder transformations, leveraging their distribution-preserving property.  Householder transformations have previously been used for other deep learning problems in very different contexts, ranging from position encoding~\cite{yang2025path, liu2025rethinking} to parameter-efficient fine-tuning~\cite{arcas2025hoft, dong2024efficient, yuan2024bridging}.

\section{Method}
\label{sec:method}

Our goal is to construct motion-embedded noise that preserves
\emph{spatial Gaussianity} while inducing 
structured temporal correlations across video frames. We achieve this through 
per-location orthogonal transformations in latent channel space, parameterized 
as products of Householder reflections.

\begin{figure}[t]
   \centering
    \includegraphics[width=\linewidth]{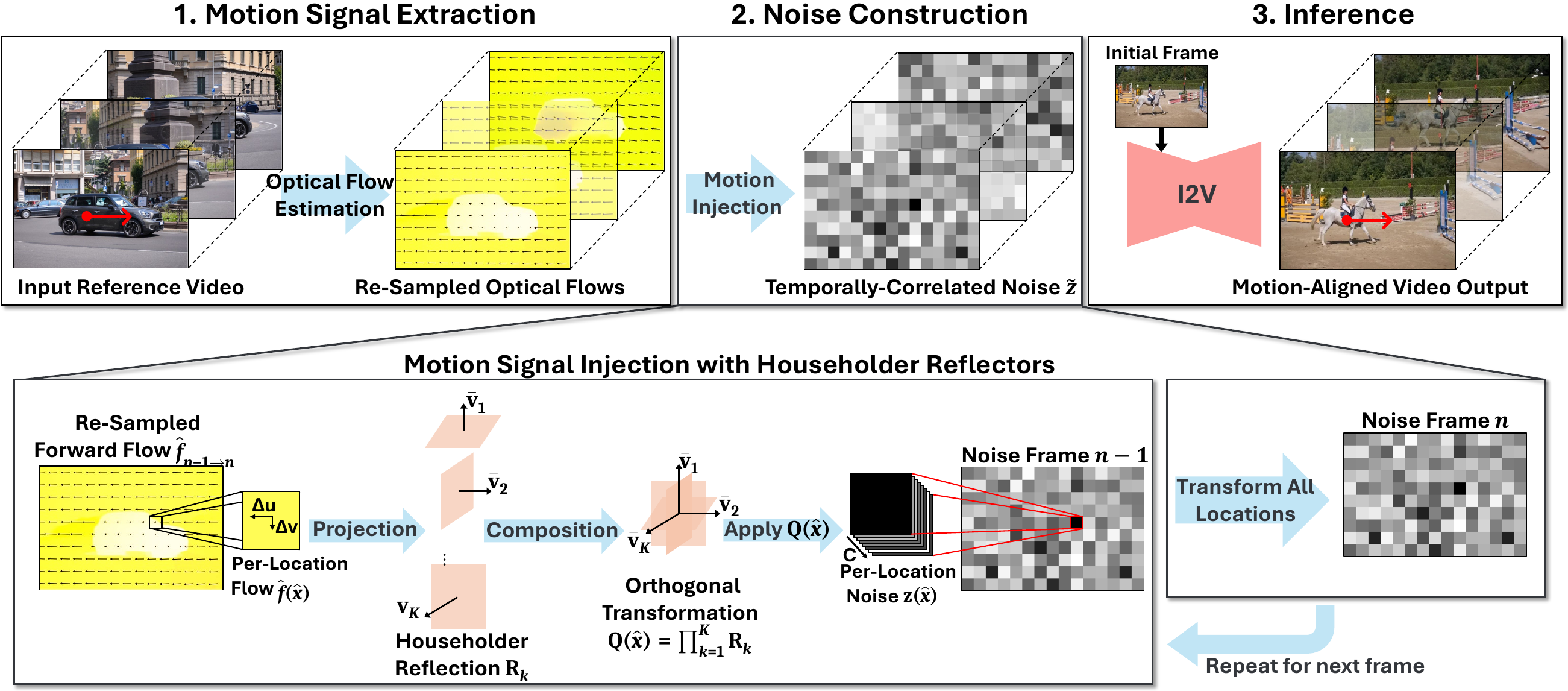}
   \vspace{-5mm}
   \caption{\textbf{HME} generates motion controlled videos in 3 stages: 1) optical flow estimation and spatial resampling, 2) temporally correlated noise construction via Householder Reflectors (see \cref{sec:householder} for details), 3) diffusion model fine-tuning and inference. At inference time, our method enables motion- and camera-controlled video generation given a text prompt, an initial frame, and a reference video as the motion signal.}
   \label{fig:pipeline}
    \vspace{-4mm}

\end{figure}

\textbf{Householder Reflector.} Householder reflectors are a standard, efficient numerical tool for constructing orthogonal transformations, requiring only an inner product and a vector update. Let $C$ denote the base model's latent channel size. A Householder reflector is 
a simple orthogonal operator defined by a single non-zero vector $\mathbf{v} \in \mathbb{R}^{C}$:
\begin{equation}
    \mathbf{R}(\mathbf{v}) = \mathbf{I}_{C} - 2\bar{\mathbf{v}}\bar{\mathbf{v}}^{\top}, 
    \qquad \bar{\mathbf{v}} = \frac{\mathbf{v}}{\|\mathbf{v}\|_{2}}.
\end{equation}
Geometrically, $\mathbf{R}(\mathbf{v})$ reflects any vector across the 
hyperplane orthogonal to $\mathbf{v}$. As reflections are orthogonal 
transformations, they preserve Euclidean norms and, crucially, the Gaussian 
distribution: if $\mathbf{z} \sim \mathcal{N}(\mathbf{0}, \mathbf{I}_{C})$, then 
$\mathbf{R}(\mathbf{v})\mathbf{z} \sim \mathcal{N}(\mathbf{0}, \mathbf{I}_{C})$ 
for any choice of $\mathbf{v}$. Composing $K$ such reflections yields a more 
expressive orthogonal operator, and any orthogonal matrix admits an exact 
decomposition into at most $C$ reflections~\cite{golub2013matrix}. We exploit 
this property to construct orthogonal transformations that
preserve spatial Gaussianity exactly.

As shown in \cref{fig:pipeline}, our framework consists of three stages: 
(1) optical flow estimation, (2) a training-free Householder noise transformation module and (3) video diffusion model 
fine-tuning. The noise module generates temporally correlated noise during both 
training and inference, while the denoising architecture and training objective 
remain unchanged.

\subsection{Motion-Embedded Noise Construction}
\label{sec:householder}

\subsubsection{Problem Formulation and Notation}
We distinguish between \emph{image space} and \emph{latent noise space}. In image space, a video $V = \{F_{0}, \ldots, F_{N-1}\}$ consists of $N$ frames $F_{n} \in \mathbb{R}^{H \times W \times 3}$ on the $H \times W$ pixel grid, and optical flow assigns each pixel $x$ a displacement $\mathbf{f}(x) = (\Delta u, \Delta v) \in \mathbb{R}^{2}$. The VAE encodes the video into $\hat{N}$ latent frames on the $\hat{H} \times \hat{W}$ latent grid and each latent location $\hat{x}$ covers a pixel footprint $\Omega(\hat{x})$ (e.g., $8 \times 8$ pixels for CogVideoX). In latent noise space, each $\hat{x}$ carries a noise vector $\mathbf{z}(\hat{x}) \in \mathbb{R}^{C}$, drawn i.i.d.\ with $\mathbf{z}(\hat{x}) \sim \mathcal{N}(\mathbf{0}, \mathbf{I}_{C})$. Operations in latent noise space act on the channel dimension only, without 
modifying spatial coordinates. In our framework, optical flow acts as a conditioning signal that parameterizes per-location orthogonal transformations of latent noise vectors, rather than spatially resampling them.

\subsubsection{Flow-to-Latent Alignment}
The optical flow between consecutive frames, $\mathbf{f}_{n-1\to n} = \Phi(F_{n-1}, F_{n})$, can be obtained with any off-the-shelf flow estimator $\Phi$; we use RAFT~\cite{teed2020raft} in our experiments. 
Since the diffusion model operates on a spatially down-sampled latent grid, 
we resample the estimated flow field to match the latent resolution:
\begin{equation}
    \hat{\mathbf{f}}(\hat{x}) = \mathrm{Resample}(\mathbf{f})(\hat{x})
    = \sum_{x \in \Omega(\hat{x})} w(x)\,\mathbf{f}_{\mathrm{norm}}(x),
    \label{eq:resample}
\end{equation}
where $\mathrm{Resample}(\cdot)$ maps the flow field from the $H \times W$ pixel grid to the 
$\hat{H} \times \hat{W}$ latent grid by computing the weighted average displacement over the 
pixel footprint $\Omega(\hat{x})$ of each latent location, and 
$\mathbf{f}_{\mathrm{norm}}(x) = (\Delta u / W,\, \Delta v / H) \in [-1, 1]^{2}$ 
denotes the flow normalized by the image resolution.

\subsubsection{Optical Flow Conditioned Parameterization}
We then convert each flow vector into $K$ transformations by applying 
fixed random linear projections:
\begin{equation}
    \mathbf{v}_{k}^{(n)}(\hat{x}) = \mathbf{A}_{k}\,\hat{\mathbf{f}}_{n-1\to n}(\hat{x}) + \mathbf{b}_{k},
    \qquad k = 1, \ldots, K.
    \label{eq:flow_proj}
\end{equation}
The flow at each latent location is two-dimensional, whereas a Householder 
reflector acting on the latent noise requires a non-zero vector in 
$\mathbb{R}^{C}$; the projections therefore lift each flow vector into $K$ 
distinct reflection directions in channel space, whose composition forms a 
$C \times C$ orthogonal transformation. 
The projections $\mathbf{A}_{k} \in \mathbb{R}^{C \times 2}$ and 
$\mathbf{b}_{k} \in \mathbb{R}^{C}$ are randomly initialized once and 
frozen throughout training and inference, introducing no learnable parameters 
and yielding a deterministic, reproducible mapping from flow to transformation. 
The only requirement on these parameters is $\mathbf{v}_{k}^{(n)}(\hat{x}) \neq \mathbf{0}$, 
since the Householder normalization is undefined for the zero vector; we use a 
small non-zero bias $\mathbf{b}_{k}$ to avoid such degenerate projections. No 
orthogonality constraint on $\mathbf{A}_{k}$ is needed, as orthogonality of the resulting 
operator is guaranteed by the Householder construction itself. Learning $\mathbf{A}_{k}$ 
and $\mathbf{b}_{k}$ is possible but unnecessary: their role is only to provide 
distinct deterministic embeddings of the flow into channel space. 
Because the projections are shared across spatial locations, the orthogonal transformation 
constructed below depends on $\hat{x}$ only through the local flow $\hat{\mathbf{f}}_{n-1\to n}(\hat{x})$. The projection parameters total only $K(2C + C) = 3KC$ 
values, independent of spatial resolution.

\subsubsection{Per-Location Orthogonal Noise Transformation}
\label{sec:ortho_transform}
The motion-embedded noise is constructed sequentially: the noise at each frame is 
obtained by applying a flow-conditioned orthogonal transformation to the 
noise of the preceding frame. Formally, for each frame $n \geq 1$ and 
latent location $\hat{x}$:
\begin{equation}
    \tilde{\mathbf{z}}_{n}(\hat{x}) = \mathbf{Q}_{n}(\hat{x})\,
    \tilde{\mathbf{z}}_{n-1}(\hat{x}),
    \qquad
    \mathbf{Q}_{n}(\hat{x}) = \prod_{k=1}^{K}
    \mathbf{R}\!\left(\mathbf{v}_{k}^{(n)}(\hat{x})\right),
    \label{eq:Q}
\end{equation}
where frame~0 retains the unmodified Gaussian noise 
$\tilde{\mathbf{z}}_{0}(\hat{x}) = \mathbf{z}_{0}(\hat{x}) \sim 
\mathcal{N}(\mathbf{0}, \mathbf{I}_{C})$, and the reflection vectors 
$\mathbf{v}_{k}^{(n)}(\hat{x})$ are obtained from the resampled forward optical flow between frames $n{-}1$ and $n$ via \cref{eq:flow_proj}.

Since each $\mathbf{Q}_{n}(\hat{x})$ is a product of Householder reflections 
and therefore orthogonal, and the product of orthogonal matrices is orthogonal, 
the noise at frame $n$ is equivalently a single orthogonal transformation of 
the initial noise:
\begin{equation}
    \tilde{\mathbf{z}}_{n}(\hat{x}) 
    = \mathbf{P}_{n}(\hat{x})\,\mathbf{z}_{0}(\hat{x}),
    \qquad
    \mathbf{P}_{n}(\hat{x}) = \mathbf{Q}_{n}(\hat{x})\cdots\mathbf{Q}_{1}(\hat{x}).
    \label{eq:unrolled}
\end{equation}
This accumulated operator $\mathbf{P}_{n}$ is orthogonal by construction, 
which is the key property exploited in Section~\ref{sec:theory}. 
The complete noise construction procedure is given in Algorithm~\ref{alg:noise}.

\paragraph{Number of Reflections and Expressivity.}
Each Householder reflection $\mathbf{R}(\mathbf{v}) = \mathbf{I}_{C} - 2\bar{\mathbf{v}}\bar{\mathbf{v}}^{\top}$ 
is a rank-one perturbation of the identity, and the composed transformation $\mathbf{Q}_{n}$ is a 
full-rank orthogonal matrix for every $K \geq 1$. 
Expanding the product in \cref{eq:Q} yields a series in the number of interacting reflections,
\begin{equation}
    \mathbf{Q}_{n}(\hat{x}) = \mathbf{I}_{C} + \sum_{m=1}^{K} (-2)^{m} \!\!\sum_{k_1 < \cdots < k_m}\!
    \Bigl(\prod_{j=1}^{m-1} \bar{\mathbf{v}}_{k_j}^{\top}\bar{\mathbf{v}}_{k_{j+1}}\Bigr)\,
    \bar{\mathbf{v}}_{k_1}\bar{\mathbf{v}}_{k_m}^{\top},
    \label{eq:expansion}
\end{equation}
where we omit the arguments $(n, \hat{x})$ of $\bar{\mathbf{v}}_{k}$. Analogous to a Taylor series, 
the order-$m$ term captures interactions among $m$ flow-dependent reflection directions and, through 
\cref{eq:flow_proj}, is a rational function of the flow $\hat{\mathbf{f}}_{n-1\to n}(\hat{x})$ of degree 
$2m$; $K$ determines the highest order retained. Since each order-$m$ term is bounded in norm by 
$2^{m}\prod_{j}|\bar{\mathbf{v}}_{k_j}^{\top}\bar{\mathbf{v}}_{k_{j+1}}|$ and the random projections 
yield reflection directions with small pairwise inner products, higher-order terms are increasingly 
attenuated and the flow influences $\mathbf{Q}_{n}$ predominantly through the low-order terms. 
Adding reflections therefore yields diminishing gains in expressivity, while the cost grows linearly 
with $K$, consistent with the saturation 
observed in \cref{tab:rank_ablation}.

{
\begin{algorithm}[t]
\caption{Motion-Embedded Noise via Householder Transformations}
\label{alg:noise}
\begin{algorithmic}[1]
    \Require Reference video $V = \{F_{0}, \ldots, F_{N-1}\}$,
             degradation level $\gamma \in [0,1]$,
             number of reflections $K$,
             fixed random projections $\{\mathbf{A}_{k}, \mathbf{b}_{k}\}_{k=1}^{K}$
    \Ensure  Motion-embedded noise tensor $\tilde{Z}_{\gamma}$

    \State $\mathbf{z}_{0}(\hat{x}) \sim \mathcal{N}(\mathbf{0}, \mathbf{I}_{C})$
           \; for all $\hat{x}$; \quad
           $\tilde{\mathbf{z}}_{0}(\hat{x}) \leftarrow \mathbf{z}_{0}(\hat{x})$

    \vspace{2pt}
    \State \textbf{Step 1: construct per frame noise}
    \For{$n = 1, \ldots, N-1$}
     
    \State \textbf{a) Estimate flows and resample}

        \State $\hat{\mathbf{f}}_{n-1 \to n}(\hat{x})
               \leftarrow \mathrm{Resample}\!\left(
               \Phi(F_{n-1}, F_{n})\right)$
        \hfill \Comment{estimate and align flow to latent grid}
    \State \textbf{b) apply orthogonal transformation to noise of previous frame}
        \State $\mathbf{v}_{k}^{(n)}(\hat{x})
               \leftarrow \mathbf{A}_{k}\,\hat{\mathbf{f}}_{n-1\to n}(\hat{x}) + \mathbf{b}_{k}$,
               \; $k = 1,\ldots,K$
        \State $\mathbf{Q}_{n}(\hat{x})
               \leftarrow \prod_{k=1}^{K} \mathbf{R}\!\left(\mathbf{v}_{k}^{(n)}(\hat{x})\right)$
               \hfill \Comment{per-location flow to orthogonal transformation}
        \State $\tilde{\mathbf{z}}_{n}(\hat{x}) 
               \leftarrow \mathbf{Q}_{n}(\hat{x})\,\tilde{\mathbf{z}}_{n-1}(\hat{x})$
    \EndFor

    \vspace{2pt}
    \State \textbf{Step 2: Stack and apply noise degradation}
    \State $\tilde{Z} \leftarrow
           \bigl[\tilde{\mathbf{z}}_{0},\,\ldots,\,\tilde{\mathbf{z}}_{N-1}\bigr]$; \quad
           $\boldsymbol{\zeta} \sim \mathcal{N}(\mathbf{0}, \mathbf{I})$, same shape as $\tilde{Z}$
    \State $\tilde{Z}_{\gamma} \leftarrow
           \dfrac{(1-\gamma)\;\tilde{Z} + \gamma\;\boldsymbol{\zeta}}
                 {\sqrt{(1-\gamma)^{2} + \gamma^{2}}}$
    \State \Return $\tilde{Z}_{\gamma}$
\end{algorithmic}
\end{algorithm}
}

\subsection{Theoretical Properties}
\label{sec:theory}
The key theoretical guarantee of our noise construction is that the transformed noise remains exactly spatially Gaussian, regardless of the motion condition; a detailed proof and empirical results are given in \cref{supp:proof}.

\begin{proposition}[Exact Spatial Gaussianity]
\label{prop:gaussianity}
Let $\tilde{\mathbf{z}}_{0}(\hat{x}) = \mathbf{z}_{0}(\hat{x}) \sim 
\mathcal{N}(\mathbf{0}, \mathbf{I}_{C})$ be i.i.d.\ across spatial locations, 
and let $\tilde{\mathbf{z}}_{n}(\hat{x}) = \mathbf{Q}_{n}(\hat{x})\,
\tilde{\mathbf{z}}_{n-1}(\hat{x})$ for $n \geq 1$, where each 
$\mathbf{Q}_{n}(\hat{x})$ is a product of Householder reflections 
parameterized by the resampled optical flow $\hat{\mathbf{f}}_{n-1 \to n}(\hat{x})$. Then 
$\tilde{\mathbf{z}}_{n}(\hat{x}) \sim \mathcal{N}(\mathbf{0}, \mathbf{I}_{C})$ 
for all $n$ and all $\hat{x}$, and the vectors $\tilde{\mathbf{z}}_{n}(\hat{x})$ 
are mutually independent across latent locations $\hat{x}$.
\end{proposition}

\paragraph{Temporal Correlation Structure.}
While \cref{prop:gaussianity} concerns each frame in isolation, the sequential construction 
couples the frames. Using \cref{eq:unrolled}, the cross-frame covariance of the noise at a 
latent location $\hat{x}$ is, for $n > m$,
\begin{equation}
    \operatorname{Cov}\!\left[\tilde{\mathbf{z}}_{n}(\hat{x}),\,\tilde{\mathbf{z}}_{m}(\hat{x})\right]
    = \mathbf{P}_{n}(\hat{x})\,\mathbf{P}_{m}(\hat{x})^{\top}
    = \mathbf{Q}_{n}(\hat{x})\cdots\mathbf{Q}_{m+1}(\hat{x}),
    \label{eq:cross_frame}
\end{equation}
while noise at different locations remains uncorrelated (\cref{supp:covariance}). 
\Cref{eq:cross_frame} shows that the dependence between two frames is an orthogonal 
$C \times C$ matrix determined entirely by the optical flow at $\hat{x}$ between frames $m$ 
and $n$: the noise vector at frame $n$ is a flow-dependent orthogonal transformation of the 
one at frame $m$. Because the projections $\{\mathbf{A}_{k}, \mathbf{b}_{k}\}$ are shared 
across locations, latent locations undergoing the same motion receive the same cross-frame 
transformation, whereas locations with different motion receive different ones, so the 
spatial pattern of these transformations mirrors the motion field. The denoising model 
(the CogVideoX DiT) is trained with this structured noise in place of 
i.i.d.\ noise and learns to exploit it as an implicit motion-conditioning signal. A geometric 
comparison with spatial noise warping is given in \cref{sec:discussion}.

\subsection{Fine-tuning with Motion-Embedded Noise}
\label{sec:training}
The key insight is that the motion conditioning is carried entirely by the cross-frame covariance structure of $\tilde{Z}$, not by its marginal distribution. Since each frame's noise is marginally standard Gaussian (\cref{prop:gaussianity}), the model trains with the standard 
diffusion objective without modification. The structured inter-frame 
dependencies, encoded by $\mathbf{Q}_n(\hat{x})$, act as an implicit 
motion signal that the denoiser learns to exploit during fine-tuning. More training details are provided in \cref{supp:training}.

\section{Experiments}
\label{sec:experiments}

\subsection{Implementation Details}
\label{sec:implementation}
We use the pretrained CogVideoX-5B-I2V as our base I2V model~\cite{yang2024cogvideox}. We fine-tune the base video diffusion model on 100K videos from RealCam-Vid~\cite{zheng2025realcam}, curated from multiple sources including RealEstate10K~\cite{zhou2018stereo}, MiraData~\cite{ju2024miradata}, and DL3DV-10K~\cite{ling2024dl3dv}, using optical flows estimated by RAFT~\cite{teed2020raft} for training. All experiments are conducted at a spatial resolution of $480\times 720$ with 49 frames per video; the latent channel dimension is $C=16$, determined by the CogVideoX VAE. We follow GWTF~\cite{burgert2025go} to fine-tune the model for 30,000 iterations using a rank-2048 LoRA setup (learning rate $1\times10^{-5}$, batch size 8), running on 4 NVIDIA A100 80GB GPUs for a total of 40 GPU-days.

\begin{figure}[t]
   \centering
    \includegraphics[width=\linewidth]{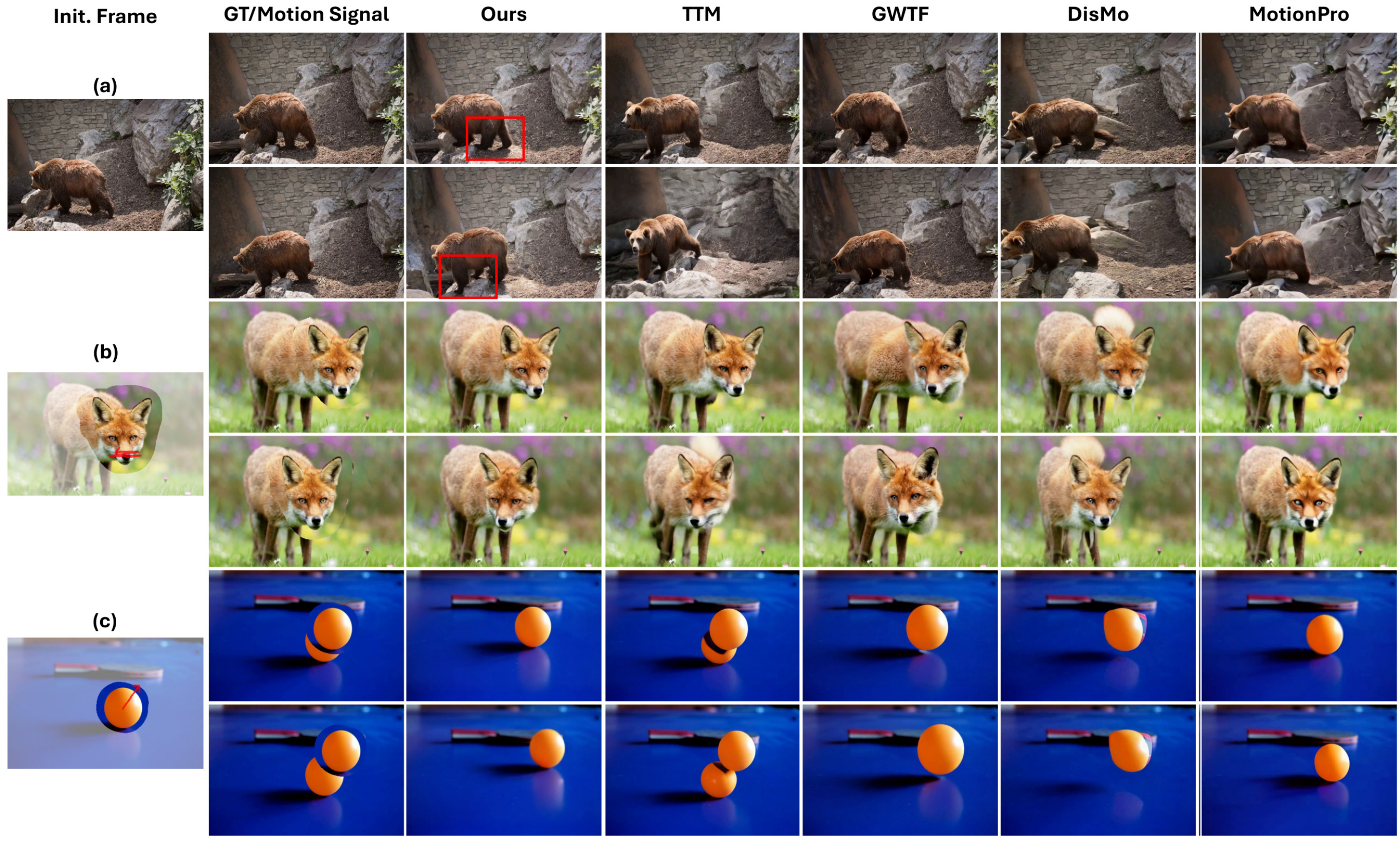}
   \vspace{-3mm}
   \caption{\textbf{Qualitative comparison of (a) motion-transfer, (b) fine-grained motion control, and (c) local object motion control I2V generation, evaluated on DAVIS and MC-Bench.} We present the intermediate frame and last frame of generated videos by each method.
   Our method demonstrates strong appearance consistency and high generation fidelity while faithfully adhering to the provided motion control signals. \textbf{Best viewed zoomed in; videos are provided in the Supplementary Material.}}
   \label{fig:Qualitative1}
   \vspace{-2mm}
\end{figure}

\begin{table*}[t]
\centering
\caption{Quantitative comparison of motion controlled I2V generation on DAVIS. Best performance in \textbf{Bold}, best baseline performance in \underline{Underline}.}
\small
\resizebox{\textwidth}{!}{%
\setlength{\tabcolsep}{4pt}

\begin{tabular}{l|cccccccccc}
\toprule
\multirow{2}{*}{Method} 
  & Subject   & BG        & Temporal  & Motion    &           & FVD                   & Optical Flow  & CoTracker & PSNR      & CLIP \\
  & Consist.\,$\uparrow$ 
             & Consist.\,$\uparrow$ 
                         & Flicker.\,$\uparrow$ 
                                     & Smooth.\,$\uparrow$ 
                                                & FID\,$\downarrow$ 
                                                            & ($\times10^3$)\,$\downarrow$ 
                                                                                   & (EPE)\,$\downarrow$ 
                                                                                               & (mIoU)\,$\uparrow$ 
                                                                                                           & (dB)\,$\uparrow$ 
                                                                                                                       & Image\,$\uparrow$ \\

\midrule
MotionClone~\cite{ling2025motionclone}        & 0.754 & 0.862 & 0.874 & 0.928 & 133.852 & 1.771 & 26.451 & 0.373 & 10.032 & 0.814 \\
ImageConductor~\cite{li2025imageconductor}      & 0.791 & 0.891 & 0.913 & 0.939 & 128.087 & 1.830 & 20.382 & 0.372 & 12.117 & 0.811 \\
GWTF$_{\gamma=0.5}$~\cite{burgert2025go} & \underline{0.888} & \underline{0.920} & \underline{0.942} & \underline{0.978} &  \underline{68.385} &  \underline{0.987} &  \underline{5.644} & 0.464 & \underline{14.748} & \underline{0.912} \\
MotionPro~\cite{zhang2025motionpro}           & 0.873 & 0.913 & 0.932 & 0.964 & 103.859 & 1.253 & 19.692 & \underline{0.473} & 14.257 & 0.879 \\
DisMo~\cite{resslerdismo} &	0.851 & 0.903 & 0.920 & 0.956 & 85.201 & 1.156 & 11.189 & 0.428 & 11.475 & 0.882 \\
TTM~\cite{singer2025time}               & 0.869 & 0.916 & 0.935 & 0.970 &  70.658 & 1.093 & 13.871 & 0.412 & 13.942 & 0.901 \\
\midrule
Ours$_{\gamma=0.1,K=2}$ & 0.896 & 0.924 & \textbf{0.951} & \textbf{0.979} &  65.343 &  \textbf{0.830} &  4.816 & 0.540 & \textbf{15.662} & 0.920 \\
Ours$_{\gamma=0.3,K=2}$& \textbf{0.898} & 0.925 & 0.946 & \textbf{0.979} &  \textbf{61.780} &  0.862 &  \textbf{4.433} & 0.546 & 15.399 & \textbf{0.923} \\
Ours$_{\gamma=0.5,K=2}$& \textbf{0.898} & \textbf{0.926} & 0.945 & \textbf{0.979} &  62.799 &  0.855 &  4.486 & \textbf{0.548} & 15.162 & \textbf{0.923} \\
\bottomrule
\end{tabular}
}
\vspace{-4mm}
\label{tab:davis}
\end{table*}

\begin{table*}[t]
\centering
\caption{Quantitative comparison of Fine-grained and Object-level motion control on MC-Bench.\\}
\label{tab:comparison}
\resizebox{\textwidth}{!}{%
\begin{tabular}{lccccccc}
\toprule
\textbf{Method} 
  & \textbf{Subject Cons.} $\uparrow$ 
  & \textbf{Background Cons.} $\uparrow$ 
  & \textbf{Motion Smooth.} $\uparrow$ 
  & \textbf{Aesthetic Qual.} $\uparrow$ 
  & \textbf{Imaging Qual.} $\uparrow$ 
  & \textbf{CTD} $\downarrow$ 
  & \textbf{Bg-Obj CTD} $\uparrow$ \\
\midrule
MotionPro~\cite{zhang2025motionpro} & 0.980	& 0.976 & \textbf{0.995} & 0.586 & 0.610 & \textbf{8.019} & 23.30 \\
TTM~\cite{singer2025time}    & \underline{0.984} & \textbf{0.981} & \textbf{0.995} & \underline{0.601} & \textbf{0.645} & \underline{8.432}  & 36.32 \\
DisMo~\cite{resslerdismo}   & \textbf{0.985} & 0.973          & 0.991          & \textbf{0.608}          & 0.601          & 
8.954           & 36.16           \\
  \midrule   
   \multicolumn{8}{l}{\textit{Noise-based methods}} \\
GWTF$_{\gamma=0.5}$~\cite{burgert2025go}    & 0.975          & 0.974          & 0.993          & 0.589          & \textbf{0.645} & 15.747          & \textbf{42.44 }          \\
Ours$_{\gamma=0.3,K=2}$ & 0.976        & \underline{0.977}          & \textbf{0.995} & 0.594          & \textbf{0.645} & 13.879           & \underline{39.53}          \\
\bottomrule
\end{tabular}%
}
\vspace{-2mm}
\end{table*}

\subsection{Video Generation with Motion Control}
\label{sec:gen}

\paragraph{Baselines and Evaluation Metrics.} We evaluate \textbf{HME} on motion transfer, fine-grained motion control, object-level motion control and camera motion control for I2V generation. The performances on DAVIS~\cite{Perazzi2016}, MC-Bench~\cite{zhang2025motionpro} are displayed in \cref{tab:davis} and \cref{tab:comparison}, respectively. Following GWTF, we adopt VBench~\cite{huang2024vbench} to assess the perceptual quality of generated video with related metrics such as Subject Consistency, Background Consistency, Temporal Flickering, and Motion Smoothness. To assess the motion adherence, we evaluate 1) CoTracker Distance and Background CoTracker distance for MC-Bench samples; 2) Optical Flow EPE, CoTracker mIoU, and PSNR against the reference videos for DAVIS samples. The input for all evaluations is a single reference image paired with a motion signal obtained from either the optical flow estimated from a reference video or derived from a user-dragged mask on the initial frame. We use the resulting forward optical flows between frames to construct the motion-embedded noise $\tilde{Z}_{\gamma}$ via~\cref{alg:noise} (details in~\cref{supp:eval}).

\paragraph{Results on Motion Transfer.} We evaluate motion transfer I2V on 45 DAVIS videos, using the motion of each original video as reference. As shown in \cref{tab:davis}, we compare against six baselines: training-free methods (MotionClone~\cite{ling2025motionclone}, TTM~\cite{singer2025time}), noise-warping (GWTF), and supervised controllers (ImageConductor~\cite{li2025imageconductor}, MotionPro~\cite{zhang2025motionpro}, DisMo~\cite{resslerdismo}). Each of our three variants ($\gamma \in \{0.1, 0.3, 0.5\}$) outperforms all six baselines on all ten metrics, despite being fine-tuned on 100K videos, compared with approximately 4M for GWTF. It attains the highest scores on all four VBench metrics, as well as the best FID and FVD. At $\gamma=0.5$, Optical Flow EPE decreases from 5.644 to 4.486, CoTracker mIoU increases from 0.464 to 0.548, and PSNR improves from 14.748 to 15.162 dB. These gains require no post-hoc Gaussianity correction, indicating that motion embedding via orthogonal transformations is a more efficient alternative to spatial noise warping. We further evaluate across varying noise degradation levels, where all variants consistently outperform GWTF and other baselines.

\vspace{-1mm}
\paragraph{Results on Fine-grained and Object-level Control.} 
We evaluate object-level and fine-grained motion control on 965 MC-Bench samples against MotionPro, TTM, DisMo, and GWTF. Our approach outperforms GWTF on five metrics, achieving lower CTD and better Background Consistency and Motion Smoothness, indicating that Householder motion embedding captures motion priors more faithfully than spatial warping. The CTD gap with supervised methods reflects a trade-off between controllability and generative freedom: TTM, MotionPro, and DisMo enforce strict trajectory adherence via explicit supervision but degrade subject-background disentanglement, evidenced by lower Bg-Obj CTD (23.30, 36.32, 36.16) relative to our method (39.53) and GWTF (42.44). Our method also matches the best Motion Smoothness (0.995) without architectural changes, 
showing that HME is competitive with supervised motion control while better preserving subject--background disentanglement.

\begin{figure}[t]
   \centering
   \vspace{-2mm}
   \includegraphics[width=\linewidth]{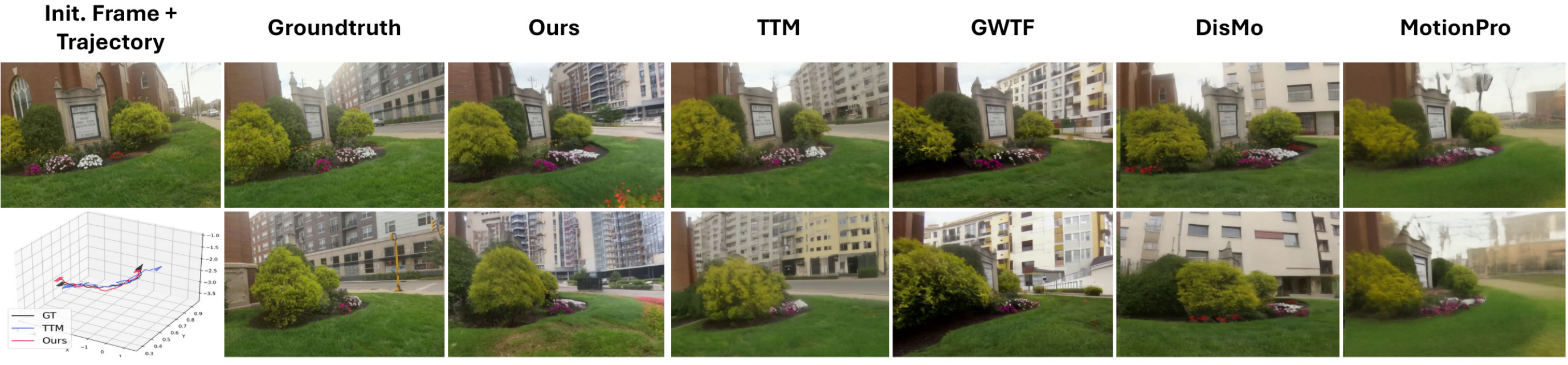}
   \caption{\textbf{Qualitative Comparison of camera motion controlled I2V generation, evaluated on DL3DV.} We present the intermediate frames and last frames of generated videos by each method. \textbf{Best viewed zoomed in; videos are provided in the Supplementary Material.}}
   \label{fig:camcontrol}
   \vspace{-3mm}
\end{figure}

\begin{table*}[t]
\centering
\caption{Quantitative comparison of DL3DV camera motion control.}
\scalebox{0.85}{%
\begin{tabular}{lccccccc}
\toprule
\textbf{Method} & \textbf{FID} $\downarrow$ & \textbf{FVD} $\downarrow$ & \textbf{PSNR (dB)} $\uparrow$ & \textbf{CLIP Image} $\uparrow$ & \textbf{LPIPS} $\downarrow$ & \textbf{SSIM} $\uparrow$ & \textbf{ATE} $\downarrow$ \\
\midrule
MotionPro~\cite{zhang2025motionpro} & 37.195 & 291.510 & 12.701 & 0.814 & 0.545 & 0.470 & 0.825 \\
DisMo~\cite{resslerdismo}           & 22.458 & \underline{187.134} & 12.255 & 0.876 & 0.506 & 0.446 & 1.155 \\
TTM~\cite{singer2025time}           & \underline{21.367} & 196.533 & \underline{12.764} & \underline{0.881} & 0.443 & 0.467 & \underline{0.464} \\
GWTF$_{\gamma=0.5}$~\cite{burgert2025go} & 23.394 & 235.260 & 12.626 & 0.880 & \underline{0.431} & \underline{0.469} & 0.473 \\
Ours$_{\gamma=0.3,K=2}$                 & \textbf{20.338} & \textbf{175.103} & \textbf{13.656} & \textbf{0.885} & \textbf{0.402} & \textbf{0.496} & \textbf{0.421} \\
\bottomrule
\end{tabular}%
}
\vspace{-4mm}
\label{tab:camera}
\end{table*}
\vspace{-2mm}
\paragraph{Results on Camera Motion Control.}
We evaluate \textbf{HME} on camera controlled I2V generation under pre-defined camera trajectories using 200 randomly sampled scenes from the DL3DV subset of the RealCam-Vid validation set. Frame-level alignment is assessed via PSNR, LPIPS, and SSIM, and camera motion consistency via Normalized Absolute Trajectory Error (ATE). As shown in \cref{tab:camera}, our method achieves the best performance across all seven metrics, outperforming noise-warping and other baselines. The leading ATE score confirms accurate replication of the target camera trajectory, while the superior PSNR, LPIPS, and SSIM scores demonstrate that our method maintains high visual fidelity alongside strong motion adherence, reflecting a great balance between appearance quality and temporal consistency under explicit camera control.

\begin{table}[t]
\centering
\vspace{-2mm}
\caption{Robustness to the flow estimator: DAVIS results with UniMatch~\cite{xu2023unifying} flows at evaluation time.}
\label{tab:unimatch}
\resizebox{\textwidth}{!}{%
\begin{tabular}{lcccccccccc}
\toprule
\multirow{2}{*}{Method} & Subject & BG & Temporal & Motion & & FVD & Optical Flow & CoTracker & PSNR & CLIP \\
 & Consist.\,$\uparrow$ & Consist.\,$\uparrow$ & Flicker.\,$\uparrow$ & Smooth.\,$\uparrow$ & FID\,$\downarrow$ & ($\times10^3$)\,$\downarrow$ & (EPE)\,$\downarrow$ & (mIoU)\,$\uparrow$ & (dB)\,$\uparrow$ & Image\,$\uparrow$ \\
\midrule
GWTF$_{\gamma=0.5}$~\cite{burgert2025go} & 0.882 & 0.918 & 0.945 & \textbf{0.979} & 73.348 & 1.180 & 5.857 & 0.472 & 14.437 & 0.901 \\
Ours$_{\gamma=0.3,K=2}$ & \textbf{0.895} & \textbf{0.926} & \textbf{0.948} & \textbf{0.979} & \textbf{63.619} & \textbf{0.837} & \textbf{4.707} & \textbf{0.541} & \textbf{15.418} & \textbf{0.915} \\
\bottomrule
\end{tabular}}
\vspace{-2mm}
\end{table}

\begin{table*}[t]
\centering
\vspace{-1mm}
\caption{Performance comparison across different numbers of Householder reflections $K$ (all with $\gamma = 0.3$), on DAVIS.}
\label{tab:rank_ablation}
\resizebox{\textwidth}{!}{
\begin{tabular}{lcccccccccc}
\toprule
\textbf{Method} 
& \textbf{Subject} $\uparrow$
& \textbf{Background} $\uparrow$
& \textbf{Temporal} $\uparrow$
& \textbf{Motion} $\uparrow$
& \textbf{FID} $\downarrow$
& \textbf{FVD} $\downarrow$
& \textbf{Flow EPE} $\downarrow$
& \textbf{CoTracker mIoU} $\uparrow$
& \textbf{PSNR} $\uparrow$
& \textbf{CLIP Image} $\uparrow$ \\
\midrule
Ours$_{K=1}$
& \textbf{0.898} 
& 0.925 
& 0.946 
& \underline{0.978} 
& 63.992 
& 861.121 
& 4.868 
& 0.543 
& 15.300 
& 0.918 \\

Ours$_{K=2}$
& \textbf{0.898} 
& 0.925 
& 0.946 
& \textbf{0.979} 
& \textbf{61.780} 
& 861.740 
& 4.433 
& 0.546 
& 15.399 
& \textbf{0.923} \\

Ours$_{K=4}$
& \textbf{0.898} 
& \underline{0.926} 
& \underline{0.948} 
& \textbf{0.979} 
& 64.579 
& \textbf{821.749} 
& 4.358 
& \underline{0.555} 
& 15.424 
& \underline{0.922} \\

Ours$_{K=8}$
& \textbf{0.898} 
& \textbf{0.927} 
& \underline{0.948} 
& \textbf{0.979} 
& \underline{63.314} 
& \underline{827.664} 
& \textbf{4.144} 
& 0.550 
& \underline{15.512} 
& \textbf{0.923} \\

Ours$_{K=12}$
& \textbf{0.898} 
& \textbf{0.927} 
& \textbf{0.950} 
& \textbf{0.979} 
& 64.699 
& 831.586 
& \underline{4.197} 
& \textbf{0.558} 
& \textbf{15.581} 
& \textbf{0.923} \\
\bottomrule
\end{tabular}
}
\vspace{-2mm}
\end{table*}

\vspace{-2mm}
\paragraph{Qualitative Results.} In \cref{fig:Qualitative1}, we present a qualitative comparison between our method and all competing baselines on (a) motion transfer, (b) fine-grained, and (c) object-level motion controlled video generation. For (a), the red box highlights that our method more faithfully replicates the detailed motion of the bear across frames, accurately capturing the displacement of the subject in accordance with the reference motion signal, whereas competing methods fail to preserve fine motions. For (b), our method not only follows the prescribed motion signal faithfully but also maintains high visual fidelity, producing coherent details throughout the sequence, while other baselines exhibit visible artifacts. For (c), our method demonstrates precise spatial awareness by correctly animating the target ball along the intended trajectory while preserving the relative positioning between objects, whereas competing methods either lose track of the motion signal or produce spatially inconsistent object placements across frames. 

We further show camera-motion-controlled I2V results in \cref{fig:camcontrol}. Compared with the strongest baseline, TTM, our generated videos yield estimated camera trajectories closer to the target trajectory, indicating stronger camera-motion adherence while better preserving viewpoint changes, scene structure, and appearance consistency. Additional results are provided in \cref{supp:extra}.

\begin{table*}[h!]
\centering
\caption{Computational overhead of noise-based methods: GPU time and approximate peak additional GPU memory on an NVIDIA RTX PRO 6000 Blackwell GPU (1024$\times$1024 resolution, $C = 16$; HIWYN with $L=3$).}
\label{tab:efficiency}
\resizebox{\textwidth}{!}{%
\begin{tabular}{lccccccccc}
\toprule
\textbf{Method} & HIWYN~\cite{chang2024warped} & GWTF~\cite{burgert2025go} & InfRes~\cite{deng2025infiniteresolution} & Ours$_{K=1}$ & Ours$_{K=2}$ & Ours$_{K=4}$ & Ours$_{K=6}$ & Ours$_{K=8}$ & Ours$_{K=12}$\\
\midrule
\textbf{GPU Time (ms/Latent) $\downarrow$}  & 38.00 & 7.63 & 23.94 & \textbf{2.27} & \underline{2.80} & 5.18 & 7.59 & 10.01 & 14.78 \\
\textbf{Memory Overhead (MB) $\downarrow$}  & 49{,}384 & 76 & 285 & \textbf{0.5} & \underline{1} & 2 & 3 & 4 & 6 \\
\bottomrule
\end{tabular}
}
\vspace{-4mm}
\end{table*}

\paragraph{Robustness to the Flow Estimator.} We replace RAFT with UniMatch~\cite{xu2023unifying} at evaluation time only, while our model remains trained on RAFT flows; GWTF and our method receive identical UniMatch flows. Our method matches or outperforms GWTF on all ten metrics (\cref{tab:unimatch}), reducing Flow EPE from $5.857$ to $4.707$.

\subsection{Ablation and Analysis}
\label{sec:ablat}
\vspace{-1mm}
\paragraph{Choice of the Number of Householder Reflections.} 

We ablate the number of Householder reflections $K \in \{1,2,4,8,12\}$ in~\cref{tab:rank_ablation}. Motion adherence is weakest at $K=1$ (Flow EPE $4.868$) and improves with diminishing returns as $K$ grows (Flow EPE $4.433 \to 4.144$ from $K=2$ to $K=8$; CoTracker mIoU $0.558$ and PSNR $15.581$\,dB at $K=12$), consistent with the attenuation of higher-order terms in~\cref{eq:expansion}, while runtime grows approximately linearly, from $2.27$\,ms/latent at $K=1$ to $14.78$\,ms/latent at $K=12$ (\cref{tab:efficiency}). At $K=2$, our method runs at $\mathbf{2.80}$\,ms/latent, a $2.7\times$ speedup over GWTF and $13.6\times$ over HIWYN. It is also memory-light: only $3KC$ projection values are stored, shared across frames and locations, and neither dense per-location $C\times C$ matrices nor noise upsampling are needed, so the peak additional memory is $0.5$--$6$\,MB, versus $76$\,MB for GWTF, $285$\,MB for InfRes and $49{,}384$\,MB for HIWYN. We use $K=2$ as the default: it achieves the best FID with only a small latency increase over $K=1$, while larger $K$ offers slightly stronger motion adherence at higher cost.
\vspace{-1mm}
\paragraph{Choice of Noise Degradation Level.} The degradation parameter $\gamma$ trades motion adherence against generative freedom. All three variants, $\gamma \in \{0.1, 0.3, 0.5\}$ at $K=2$, outperform GWTF on all ten metrics (\cref{tab:davis}); lower $\gamma$ strengthens motion adherence, while higher $\gamma$ favours perceptual quality and subject consistency, with small differences overall. We adopt $\gamma=0.3$ for all other experiments; qualitative comparisons are in~\cref{supp:ablat}.

\vspace{-0.1in}
\section{Discussion}
\label{sec:discussion}
\paragraph{Geometric Analysis of Temporal Noise Correlation.}
PYoCo~\cite{ge2023pyoco} shows that correlating noise across frames promotes temporal coherence. GWTF~\cite{burgert2025go} realises this by transporting noise along flow trajectories \emph{across} locations, whereas our noise stays tied to its location: $\tilde{\mathbf{z}}_{n}(\hat{x}) = \mathbf{P}_{n}(\hat{x})\,\mathbf{z}_{0}(\hat{x})$ preserves its norm, and its change between frames, $(\mathbf{Q}_{n}(\hat{x}) - \mathbf{I}_{C})\,\tilde{\mathbf{z}}_{n-1}(\hat{x})$, depends only on the local flow. Since the projections are shared across locations, regions with the same motion receive the same transformation; this includes static regions, since $\mathbf{Q}_{n} \neq \mathbf{I}_{C}$ for zero flow when $\mathbf{b}_{k} \neq \mathbf{0}$. We hypothesise that this per-location encoding of motion in channel space provides a more direct conditioning signal than spatial transport, contributing to the improved motion adherence over GWTF.

\paragraph{Limitations.} Despite its effectiveness, our method has several limitations. The choice of the number of reflections $K$ introduces an expressivity-efficiency trade-off and may require tuning for different settings. The method also relies on accurate optical flow for training, so noisy motion estimates can weaken motion adherence. Relatedly, since the motion signal is carried by per-location flow, subjects undergoing complex non-rigid geometric transformations (e.g., the rapidly rotating snowboarder in DAVIS) remain challenging, which we leave for future work. Moreover, while the noise construction is training-free and parameter-free, the video diffusion model still requires fine-tuning to fully exploit the motion-embedded noise. Finally, the baselines are evaluated with their released checkpoints rather than retrained on our training data and RAFT flows. 

\vspace{-0.1in}\section{Conclusion}\vspace{-0.1in}
We presented \textbf{Householder Motion Embedding (HME)}, a noise-modulation framework for motion-controllable image-to-video generation. By replacing spatial noise transport with per-location Householder orthogonal transformations in latent channel space, our method preserves spatial Gaussianity by construction while encoding motion-aligned temporal correlations. Experiments on motion transfer, object-level control, and camera motion control show that \textbf{HME} improves motion adherence, visual fidelity, and noise-processing efficiency over existing noise-warping methods. Beyond noise construction, we believe that geometric transformation-based conditioning offers a promising way for incorporating structured motion priors into video diffusion models.
\vspace{-2mm}

\begin{ack}
This work was conducted in collaboration with Vertex Lab Technology Limited. This research is funded in part by an Australian Research Council (ARC) Discovery Grant DP220100800.  \mbox{Hongdong Li} holds concurrent appointments with The Australian National University (full-time) and with Amazon as an Amazon Scholar (part-time).
\end{ack}

\bibliographystyle{plainnat}
\bibliography{mybib}

\begin{thebibliography}{68}
\providecommand{\natexlab}[1]{#1}
\providecommand{\url}[1]{\texttt{#1}}
\expandafter\ifx\csname urlstyle\endcsname\relax
  \providecommand{\doi}[1]{doi: #1}\else
  \providecommand{\doi}{doi: \begingroup \urlstyle{rm}\Url}\fi

\bibitem[Aira et~al.(2024)Aira, Montanaro, Aiello, Valsesia, and
  Magli]{aira2024motioncraft}
Luca~Savant Aira, Antonio Montanaro, Emanuele Aiello, Diego Valsesia, and
  Enrico Magli.
\newblock Motioncraft: Physics-based zero-shot video generation.
\newblock In \emph{NeurIPS}, 2024.

\bibitem[Arcas et~al.(2025)Arcas, Sanchis, Civera, and Juan]{arcas2025hoft}
Alejandro~Moreno Arcas, Albert Sanchis, Jorge Civera, and Alfons Juan.
\newblock Hoft: Householder orthogonal fine-tuning.
\newblock \emph{arXiv preprint arXiv:2505.16531}, 2025.

\bibitem[Bahmani et~al.(2025)Bahmani, Skorokhodov, Siarohin, Menapace, Qian,
  Vasilkovsky, Lee, Wang, Zou, Tagliasacchi, Lindell, and
  Tulyakov]{bahmani2025vd3d}
Sherwin Bahmani, Ivan Skorokhodov, Aliaksandr Siarohin, Willi Menapace,
  Guocheng Qian, Michael Vasilkovsky, Hsin-Ying Lee, Chaoyang Wang, Jiaxu Zou,
  Andrea Tagliasacchi, David~B. Lindell, and Sergey Tulyakov.
\newblock {VD3D}: Taming large video diffusion transformers for {3D} camera
  control.
\newblock In \emph{ICLR}, 2025.
\newblock URL \url{https://openreview.net/forum?id=0n4bS0R5MM}.

\bibitem[Blattmann et~al.(2023{\natexlab{a}})Blattmann, Dockhorn, Kulal,
  Mendelevitch, Kilian, Lorenz, Levi, English, Voleti, Letts,
  et~al.]{blattmann2023stable}
Andreas Blattmann, Tim Dockhorn, Sumith Kulal, Daniel Mendelevitch, Maciej
  Kilian, Dominik Lorenz, Yam Levi, Zion English, Vikram Voleti, Adam Letts,
  et~al.
\newblock Stable video diffusion: Scaling latent video diffusion models to
  large datasets.
\newblock \emph{arXiv preprint arXiv:2311.15127}, 2023{\natexlab{a}}.

\bibitem[Blattmann et~al.(2023{\natexlab{b}})Blattmann, Rombach, Ling,
  Dockhorn, Kim, Fidler, and Kreis]{blattmann2023align}
Andreas Blattmann, Robin Rombach, Huan Ling, Tim Dockhorn, Seung~Wook Kim,
  Sanja Fidler, and Karsten Kreis.
\newblock {Align your Latents: High-Resolution Video Synthesis with Latent
  Diffusion Models}.
\newblock In \emph{CVPR}, 2023{\natexlab{b}}.

\bibitem[Brooks et~al.(2024)Brooks, Peebles, Holmes, DePue, Guo, Jing, Schnurr,
  Taylor, Luhman, Luhman, Ng, Wang, and Ramesh]{videoworldsimulators2024}
Tim Brooks, Bill Peebles, Connor Holmes, Will DePue, Yufei Guo, Li~Jing, David
  Schnurr, Joe Taylor, Troy Luhman, Eric Luhman, Clarence Ng, Ricky Wang, and
  Aditya Ramesh.
\newblock Video generation models as world simulators.
\newblock OpenAI technical report, 2024.
\newblock URL
  \url{https://openai.com/research/video-generation-models-as-world-simulators}.

\bibitem[Burgert et~al.(2025)Burgert, Xu, Xian, Pilarski, Clausen, He, Ma,
  Deng, Li, Mousavi, et~al.]{burgert2025go}
Ryan Burgert, Yuancheng Xu, Wenqi Xian, Oliver Pilarski, Pascal Clausen,
  Mingming He, Li~Ma, Yitong Deng, Lingxiao Li, Mohsen Mousavi, et~al.
\newblock Go-with-the-flow: Motion-controllable video diffusion models using
  real-time warped noise.
\newblock In \emph{CVPR}, pages 13--23, 2025.

\bibitem[Chang et~al.(2024)Chang, Tang, Gross, and Azevedo]{chang2024warped}
Pascal Chang, Jingwei Tang, Markus Gross, and Vinicius~C Azevedo.
\newblock How i warped your noise: a temporally-correlated noise prior for
  diffusion models.
\newblock In \emph{ICLR}. OpenReview, 2024.

\bibitem[Chen et~al.(2023)Chen, Ji, Wu, Wu, Xie, Li, Xia, Xiao, and
  Lin]{zhang2024controlvideo2}
Weifeng Chen, Yatai Ji, Jie Wu, Hefeng Wu, Pan Xie, Jiashi Li, Xin Xia, Xuefeng
  Xiao, and Liang Lin.
\newblock {Control-A-Video: Controllable Text-to-Video Diffusion Models with
  Motion Prior and Reward Feedback Learning}.
\newblock \emph{arXiv preprint arXiv:2305.13840}, 2023.

\bibitem[Deng et~al.(2025)Deng, Lin, Li, Smirnov, Burgert, Yu, Dedun, and
  Taghavi]{deng2025infiniteresolution}
Yitong Deng, Winnie Lin, Lingxiao Li, Dmitriy Smirnov, Ryan~D Burgert, Ning Yu,
  Vincent Dedun, and Mohammad~H. Taghavi.
\newblock Infinite-resolution integral noise warping for diffusion models.
\newblock In \emph{ICLR}, 2025.
\newblock URL \url{https://openreview.net/forum?id=Y6LPWBo2HP}.

\bibitem[Dong et~al.(2024)Dong, Sun, Yang, Zhang, Lin, Yan, Zhang, Wang, Yang,
  and Shen]{dong2024efficient}
Wei Dong, Yuan Sun, Yiting Yang, Xing Zhang, Zhijun Lin, Qingsen Yan, Haokui
  Zhang, Peng Wang, Yang Yang, and Hengtao Shen.
\newblock Efficient adaptation of pre-trained vision transformer via
  householder transformation.
\newblock \emph{NeurIPS}, 37:\penalty0 102056--102077, 2024.

\bibitem[Gao et~al.(2025)Gao, Guo, Hoang, Huang, Jiang, Kong, Li, Li, Li, Li,
  et~al.]{gao2025seedance}
Yu~Gao, Haoyuan Guo, Tuyen Hoang, Weilin Huang, Lu~Jiang, Fangyuan Kong, Huixia
  Li, Jiashi Li, Liang Li, Xiaojie Li, et~al.
\newblock Seedance 1.0: Exploring the boundaries of video generation models.
\newblock \emph{arXiv preprint arXiv:2506.09113}, 2025.

\bibitem[Ge et~al.(2023)Ge, Nah, Liu, Poon, Tao, Catanzaro, Jacobs, Huang, Liu,
  and Balaji]{ge2023pyoco}
Songwei Ge, Seungjun Nah, Guilin Liu, Tyler Poon, Andrew Tao, Bryan Catanzaro,
  David Jacobs, Jia-Bin Huang, Ming-Yu Liu, and Yogesh Balaji.
\newblock {Preserve Your Own Correlation: A Noise Prior for Video Diffusion
  Models}.
\newblock In \emph{ICCV}, 2023.

\bibitem[Geng et~al.(2025)Geng, Herrmann, Hur, Cole, Zhang, Pfaff,
  Lopez-Guevara, Aytar, Rubinstein, Sun, et~al.]{geng2025motion}
Daniel Geng, Charles Herrmann, Junhwa Hur, Forrester Cole, Serena Zhang, Tobias
  Pfaff, Tatiana Lopez-Guevara, Yusuf Aytar, Michael Rubinstein, Chen Sun,
  et~al.
\newblock Motion prompting: Controlling video generation with motion
  trajectories.
\newblock In \emph{CVPR}, pages 1--12, 2025.

\bibitem[Geyer et~al.(2024)Geyer, Bar-Tal, Bagon, and
  Dekel]{geyer2024tokenflow}
Michal Geyer, Omer Bar-Tal, Shai Bagon, and Tali Dekel.
\newblock Tokenflow: Consistent diffusion features for consistent video
  editing.
\newblock In \emph{ICLR}, 2024.
\newblock URL \url{https://arxiv.org/abs/2307.10373}.

\bibitem[G{\"o}kmen et~al.(2025)G{\"o}kmen, Ekin, Bilecen, and
  Dundar]{gokmen2025ropecraft}
Ahmet~Berke G{\"o}kmen, Yi{\u{g}}it Ekin, Bahri~Batuhan Bilecen, and Aysegul
  Dundar.
\newblock Ropecraft: Training-free motion transfer with trajectory-guided rope
  optimization on diffusion transformers.
\newblock In \emph{NeurIPS}, 2025.

\bibitem[Golub and Van~Loan(2013)]{golub2013matrix}
Gene~H Golub and Charles~F Van~Loan.
\newblock \emph{Matrix computations}.
\newblock JHU press, 2013.

\bibitem[Gu et~al.(2025)]{gu2025diffusion2}
Zekai Gu et~al.
\newblock Diffusion as shader.
\newblock In \emph{SIGGRAPH}, 2025.

\bibitem[Guo et~al.(2024)Guo, Yang, Rao, Liang, Wang, Qiao, Agrawala, Lin, and
  Dai]{guo2024animatediff}
Yuwei Guo, Ceyuan Yang, Anyi Rao, Zhengyang Liang, Yaohui Wang, Yu~Qiao,
  Maneesh Agrawala, Dahua Lin, and Bo~Dai.
\newblock Animatediff: Animate your personalized text-to-image diffusion models
  without specific tuning.
\newblock In \emph{ICLR}, 2024.

\bibitem[He et~al.(2025)He, Xu, Guo, Wetzstein, Dai, Li, and
  Yang]{he2024cameractrl}
Hao He, Yinghao Xu, Yuwei Guo, Gordon Wetzstein, Bo~Dai, Hongsheng Li, and
  Ceyuan Yang.
\newblock {CameraCtrl}: Enabling camera control for video diffusion models.
\newblock In \emph{ICLR}, 2025.

\bibitem[Ho et~al.(2022)Ho, Salimans, Gritsenko, Chan, Norouzi, and
  Fleet]{ho2022vdm}
Jonathan Ho, Tim Salimans, Alexey Gritsenko, William Chan, Mohammad Norouzi,
  and David~J. Fleet.
\newblock {Video Diffusion Models}.
\newblock In \emph{NeurIPS}, 2022.

\bibitem[Hoogeboom et~al.(2019)Hoogeboom, van~den Berg, and
  Welling]{hoogeboom2019emerging}
Emiel Hoogeboom, Rianne van~den Berg, and Max Welling.
\newblock Emerging convolutions for generative normalizing flows.
\newblock In \emph{ICML}, pages 2771--2780. PMLR, 2019.

\bibitem[Huang et~al.(2024)Huang, He, Yu, Zhang, Si, Jiang, Zhang, Wu, Jin,
  Chanpaisit, et~al.]{huang2024vbench}
Ziqi Huang, Yinan He, Jiashuo Yu, Fan Zhang, Chenyang Si, Yuming Jiang, Yuanhan
  Zhang, Tianxing Wu, Qingyang Jin, Nattapol Chanpaisit, et~al.
\newblock Vbench: Comprehensive benchmark suite for video generative models.
\newblock In \emph{CVPR}, 2024.

\bibitem[Jain et~al.(2024)Jain, Nasery, Vineet, and Behl]{jain2024peekaboo}
Yash Jain, Anshul Nasery, Vibhav Vineet, and Harkirat Behl.
\newblock Peekaboo: Interactive video generation via masked-diffusion.
\newblock In \emph{CVPR}, 2024.

\bibitem[Ju et~al.(2024)Ju, Gao, Zhang, Yuan, Wang, Zeng, Xiong, Xu, and
  Shan]{ju2024miradata}
Xuan Ju, Yiming Gao, Zhaoyang Zhang, Ziyang Yuan, Xintao Wang, Ailing Zeng,
  Yu~Xiong, Qiang Xu, and Ying Shan.
\newblock Miradata: A large-scale video dataset with long durations and
  structured captions.
\newblock \emph{NeurIPS}, 37:\penalty0 48955--48970, 2024.

\bibitem[Karaev et~al.(2024)Karaev, Rocco, Graham, Neverova, Vedaldi, and
  Rupprecht]{karaev2024cotracker}
Nikita Karaev, Ignacio Rocco, Benjamin Graham, Natalia Neverova, Andrea
  Vedaldi, and Christian Rupprecht.
\newblock Cotracker: It is better to track together.
\newblock In \emph{ECCV}, 2024.
\newblock URL \url{https://arxiv.org/abs/2307.07635}.

\bibitem[Kong et~al.(2024)Kong, Tian, Zhang, Min, Dai, Zhou, Xiong, Li, Wu,
  Zhang, et~al.]{kong2024hunyuanvideo}
Weijie Kong, Qi~Tian, Zijian Zhang, Rox Min, Zuozhuo Dai, Jin Zhou, Jiangfeng
  Xiong, Xin Li, Bo~Wu, Jianwei Zhang, et~al.
\newblock Hunyuanvideo: A systematic framework for large video generative
  models.
\newblock \emph{arXiv preprint arXiv:2412.03603}, 2024.

\bibitem[Ku et~al.(2024)Ku, Wei, Ren, Yang, and Chen]{ku2024anyv2v}
Max Ku, Cong Wei, Weiming Ren, Huan Yang, and Wenhu Chen.
\newblock {AnyV2V}: A tuning-free framework for any video-to-video editing
  tasks.
\newblock \emph{TMLR}, 2024.

\bibitem[Kuang et~al.(2024)Kuang, Cai, He, Xu, Li, Guibas, and
  Wetzstein]{kuang2024collaborative}
Zhengfei Kuang, Shengqu Cai, Hao He, Yinghao Xu, Hongsheng Li, Leonidas Guibas,
  and Gordon Wetzstein.
\newblock Collaborative video diffusion: Consistent multi-video generation with
  camera control.
\newblock In \emph{NeurIPS}, 2024.

\bibitem[Li et~al.(2025)Li, Wang, Zhang, Wang, Yuan, Xie, Shan, and
  Zou]{li2025imageconductor}
Yaowei Li, Xintao Wang, Zhaoyang Zhang, Zhouxia Wang, Ziyang Yuan, Liangbin
  Xie, Ying Shan, and Yuexian Zou.
\newblock Image conductor: Precision control for interactive video synthesis.
\newblock In \emph{AAAI}, volume~39, pages 5031--5038, 2025.

\bibitem[Ling et~al.(2024)Ling, Sheng, Tu, Zhao, Xin, Wan, Yu, Guo, Yu, Lu,
  et~al.]{ling2024dl3dv}
Lu~Ling, Yichen Sheng, Zhi Tu, Wentian Zhao, Cheng Xin, Kun Wan, Lantao Yu,
  Qianyu Guo, Zixun Yu, Yawen Lu, et~al.
\newblock Dl3dv-10k: A large-scale scene dataset for deep learning-based 3d
  vision.
\newblock In \emph{CVPR}, pages 22160--22169, 2024.

\bibitem[Ling et~al.(2025)Ling, Bu, Zhang, Dong, Zang, Wu, Chen, Wang, and
  Jin]{ling2025motionclone}
Pengyang Ling, Jiazi Bu, Pan Zhang, Xiaoyi Dong, Yuhang Zang, Tong Wu, Huaian
  Chen, Jiaqi Wang, and Yi~Jin.
\newblock {MotionClone}: Training-free motion cloning for controllable video
  generation.
\newblock In \emph{ICLR}, 2025.
\newblock URL \url{https://openreview.net/forum?id=aY3L65HgHJ}.

\bibitem[Liu et~al.(2025)Liu, Lin, Sun, Shangguan, Alvarez, and
  Zhou]{liu2025rethinking}
Haiping Liu, Lijing Lin, Jingyuan Sun, Zhegong Shangguan, Mauricio~A Alvarez,
  and Hongpeng Zhou.
\newblock Rethinking rope: A mathematical blueprint for n-dimensional
  positional embedding.
\newblock \emph{arXiv preprint arXiv:2504.06308}, 2025.

\bibitem[Mhammedi et~al.(2017)Mhammedi, Hellicar, Rahman, and
  Bailey]{mhammedi2017efficient}
Zakaria Mhammedi, Andrew Hellicar, Ashfaqur Rahman, and James Bailey.
\newblock Efficient orthogonal parametrisation of recurrent neural networks
  using householder reflections.
\newblock In \emph{ICML}, pages 2401--2409. PMLR, 2017.

\bibitem[Mou et~al.(2024)Mou, Cao, Wang, Zhang, Shan, and
  Zhang]{mou2024revideo}
Chong Mou, Mingdeng Cao, Xintao Wang, Zhaoyang Zhang, Ying Shan, and Jian
  Zhang.
\newblock Revideo: Remake a video with motion and content control.
\newblock \emph{arXiv}, 2024.

\bibitem[Namekata et~al.(2025)Namekata, Bahmani, Wu, Kant, Gilitschenski, and
  Lindell]{namekata2025sgi2v}
Koichi Namekata, Sherwin Bahmani, Ziyi Wu, Yash Kant, Igor Gilitschenski, and
  David~B. Lindell.
\newblock {SG-I2V}: Self-guided trajectory control in image-to-video
  generation.
\newblock In \emph{ICLR}, 2025.
\newblock URL \url{https://openreview.net/forum?id=uQjySppU9x}.

\bibitem[Niu et~al.(2024)Niu, Cun, Wang, Zhang, Shan, and Zheng]{niu2024mofa}
Muyao Niu, Xiaodong Cun, Xintao Wang, Yong Zhang, Ying Shan, and Yinqiang
  Zheng.
\newblock {MOFA-Video: Controllable Image Animation via Generative Motion Field
  Adaptions in Frozen Image-to-Video Diffusion Model}.
\newblock In \emph{ECCV}, 2024.

\bibitem[Peebles and Xie(2023)]{peebles2023scalable}
William Peebles and Saining Xie.
\newblock Scalable diffusion models with transformers.
\newblock In \emph{ICCV}, 2023.

\bibitem[Perazzi et~al.(2016)Perazzi, Pont-Tuset, McWilliams, {Van Gool},
  Gross, and Sorkine-Hornung]{Perazzi2016}
F.~Perazzi, J.~Pont-Tuset, B.~McWilliams, L.~{Van Gool}, M.~Gross, and
  A.~Sorkine-Hornung.
\newblock A benchmark dataset and evaluation methodology for video object
  segmentation.
\newblock In \emph{CVPR}, 2016.

\bibitem[Pondaven et~al.(2025)Pondaven, Siarohin, Tulyakov, Torr, and
  Pizzati]{pondaven2025ditflow}
Alexander Pondaven, Aliaksandr Siarohin, Sergey Tulyakov, Philip Torr, and
  Fabio Pizzati.
\newblock Video motion transfer with diffusion transformers.
\newblock In \emph{CVPR}, 2025.

\bibitem[Qiu et~al.(2023)Qiu, Liu, Feng, Xue, Feng, Liu, Zhang, Weller, and
  Sch{\"o}lkopf]{qiu2023controlling}
Zeju Qiu, Weiyang Liu, Haiwen Feng, Yuxuan Xue, Yao Feng, Zhen Liu, Dan Zhang,
  Adrian Weller, and Bernhard Sch{\"o}lkopf.
\newblock Controlling text-to-image diffusion by orthogonal finetuning.
\newblock \emph{NeurIPS}, 36:\penalty0 79320--79362, 2023.

\bibitem[Ren et~al.(2025)Ren, Shen, Huang, Ling, Lu, Nimier-David, M{\"u}ller,
  Keller, Fidler, and Gao]{ren2025gen3c}
Xuanchi Ren, Tianchang Shen, Jiahui Huang, Huan Ling, Yifan Lu, Merlin
  Nimier-David, Thomas M{\"u}ller, Alexander Keller, Sanja Fidler, and Jun Gao.
\newblock Gen3c: 3d-informed world-consistent video generation with precise
  camera control.
\newblock In \emph{CVPR}, pages 6121--6132, 2025.

\bibitem[Ressler-Antal et~al.(2025)Ressler-Antal, Fundel, Alaya, Baumann,
  Krause, Gui, and Ommer]{resslerdismo}
Thomas Ressler-Antal, Frank Fundel, Malek~Ben Alaya, Stefan~Andreas Baumann,
  Felix Krause, Ming Gui, and Bj{\"o}rn Ommer.
\newblock Dismo: Disentangled motion representations for open-world motion
  transfer.
\newblock In \emph{NeurIPS}, 2025.

\bibitem[Sch{\"o}nberger and Frahm(2016)]{schoenberger2016sfm}
Johannes~Lutz Sch{\"o}nberger and Jan-Michael Frahm.
\newblock Structure-from-motion revisited.
\newblock In \emph{CVPR}, 2016.

\bibitem[Shi et~al.(2024)Shi, Huang, Wang, Bian, Li, Zhang, Meng, Cheung, See,
  Qin, Dai, and Li]{shi2024motioni2v}
Xiaoyu Shi, Zhaoyang Huang, Fu-Yun Wang, Weikang Bian, Dasong Li, Yi~Zhang,
  Manyuan Meng, Ka~Chun Cheung, Simon See, Hongwei Qin, Jifeng Dai, and
  Hongsheng Li.
\newblock Motion-i2v: Consistent and controllable image-to-video generation
  with explicit motion modeling.
\newblock In \emph{SIGGRAPH}, 2024.
\newblock URL \url{https://arxiv.org/abs/2401.15977}.

\bibitem[Singer et~al.(2026)Singer, Rotstein, Mann, Kimmel, and
  Litany]{singer2025time}
Assaf Singer, Noam Rotstein, Amir Mann, Ron Kimmel, and Or~Litany.
\newblock {Time-to-Move}: Training-free motion-controlled video generation via
  dual-clock denoising.
\newblock In \emph{ICLR}, 2026.

\bibitem[Teed and Deng(2020)]{teed2020raft}
Zachary Teed and Jia Deng.
\newblock {RAFT}: Recurrent all-pairs field transforms for optical flow.
\newblock In \emph{ECCV}, pages 402--419, 2020.
\newblock URL \url{https://arxiv.org/abs/2003.12039}.

\bibitem[Tomczak and Welling(2016)]{tomczak2016improving}
Jakub~M Tomczak and Max Welling.
\newblock Improving variational auto-encoders using householder flow.
\newblock \emph{arXiv preprint arXiv:1611.09630}, 2016.

\bibitem[Wan et~al.(2025)Wan, Wang, Ai, Wen, Mao, Xie, Chen, Yu, Zhao, Yang,
  et~al.]{wan2025wan}
Team Wan, Ang Wang, Baole Ai, Bin Wen, Chaojie Mao, Chen-Wei Xie, Di~Chen,
  Feiwu Yu, Haiming Zhao, Jianxiao Yang, et~al.
\newblock Wan: Open and advanced large-scale video generative models.
\newblock \emph{arXiv preprint arXiv:2503.20314}, 2025.

\bibitem[Wang et~al.(2024{\natexlab{a}})Wang, Yuan, Zhang, Chen, Wang, Zhang,
  Shen, Zhao, and Zhou]{wang2024videocomposer}
Xiang Wang, Hangjie Yuan, Shiwei Zhang, Dayou Chen, Jiuniu Wang, Yingya Zhang,
  Yujun Shen, Deli Zhao, and Jingren Zhou.
\newblock Videocomposer: Compositional video synthesis with motion
  controllability.
\newblock In \emph{NeurIPS}, 2024{\natexlab{a}}.

\bibitem[Wang et~al.(2026)Wang, Zhang, Cai, Wu, Ackermann, Kuang, Zheng,
  Raji{\v{c}}, Tang, and Wetzstein]{wang2025bullettime}
Yiming Wang, Qihang Zhang, Shengqu Cai, Tong Wu, Jan Ackermann, Zhengfei Kuang,
  Yang Zheng, Frano Raji{\v{c}}, Siyu Tang, and Gordon Wetzstein.
\newblock {BulletTime}: Decoupled control of time and camera pose for video
  generation.
\newblock In \emph{CVPR}, 2026.

\bibitem[Wang et~al.(2024{\natexlab{b}})Wang, Yuan, Wang, Li, Chen, Xia, Luo,
  and Shan]{wang2024motionctrl}
Zhouxia Wang, Ziyang Yuan, Xintao Wang, Yaowei Li, Tianshui Chen, Menghan Xia,
  Ping Luo, and Ying Shan.
\newblock {MotionCtrl}: A unified and flexible motion controller for video
  generation.
\newblock In \emph{SIGGRAPH}, pages 1--11, 2024{\natexlab{b}}.

\bibitem[Wu et~al.(2024)Wu, Li, Gu, Zhao, He, Zhang, Shou, Li, Gao, and
  Zhang]{wu2024draganything}
Weijia Wu, Zhuang Li, Yuchao Gu, Rui Zhao, Yefei He, David~Junhao Zhang,
  Mike~Zheng Shou, Yan Li, Tingting Gao, and Di~Zhang.
\newblock {DragAnything}: Motion control for anything using entity
  representation.
\newblock In \emph{ECCV}, 2024.

\bibitem[Xing et~al.(2024)Xing, Xia, Zhang, Chen, Wang, Wong, and
  Shan]{xing2024dynamicrafter}
Jinbo Xing, Menghan Xia, Yong Zhang, Haoxin Chen, Xintao Wang, Tien-Tsin Wong,
  and Ying Shan.
\newblock {DynamiCrafter: Animating Open-domain Images with Video Diffusion
  Priors}.
\newblock In \emph{ECCV}, 2024.

\bibitem[Xu et~al.(2024)Xu, Nie, Liu, Liu, Kautz, Wang, and
  Vahdat]{xu2024camco}
Dejia Xu, Weili Nie, Chao Liu, Sifei Liu, Jan Kautz, Zhangyang Wang, and Arash
  Vahdat.
\newblock Camco: Camera-controllable 3d-consistent image-to-video generation.
\newblock \emph{arXiv preprint arXiv:2406.02509}, 2024.

\bibitem[Xu et~al.(2023)Xu, Zhang, Cai, Rezatofighi, Yu, Tao, and
  Geiger]{xu2023unifying}
Haofei Xu, Jing Zhang, Jianfei Cai, Hamid Rezatofighi, Fisher Yu, Dacheng Tao,
  and Andreas Geiger.
\newblock Unifying flow, stereo and depth estimation.
\newblock \emph{TPAMI}, 45\penalty0 (11):\penalty0 13941--13958, 2023.

\bibitem[Yang et~al.(2024)Yang, Hou, Huang, Ma, Wan, Zhang, Chen, and
  Liao]{yang2024direct}
Shiyuan Yang, Liang Hou, Haibin Huang, Chongyang Ma, Pengfei Wan, Di~Zhang,
  Xiaodong Chen, and Jing Liao.
\newblock Direct-a-video: Customized video generation with user-directed camera
  movement and object motion.
\newblock In \emph{SIGGRAPH}, 2024.

\bibitem[Yang et~al.(2025{\natexlab{a}})Yang, Shen, Wen, Tan, Mishra, Ren,
  Panda, and Kim]{yang2025path}
Songlin Yang, Yikang Shen, Kaiyue Wen, Shawn Tan, Mayank Mishra, Liliang Ren,
  Rameswar Panda, and Yoon Kim.
\newblock Path attention: Position encoding via accumulating householder
  transformations.
\newblock In \emph{NeurIPS}, 2025{\natexlab{a}}.

\bibitem[Yang et~al.(2025{\natexlab{b}})Yang, Teng, Zheng, Ding, Huang, Xu,
  Yang, Hong, Zhang, Feng, Dong, et~al.]{yang2024cogvideox}
Zhuoyi Yang, Jiayan Teng, Wendi Zheng, Ming Ding, Shiyu Huang, Jiazheng Xu,
  Yuanming Yang, Wenyi Hong, Xiaohan Zhang, Guanyu Feng, Da~Dong, et~al.
\newblock {CogVideoX}: Text-to-video diffusion models with an expert
  transformer.
\newblock In \emph{ICLR}, 2025{\natexlab{b}}.
\newblock URL \url{https://arxiv.org/abs/2408.06072}.

\bibitem[Yatim et~al.(2024)Yatim, Fridman, Bar-Tal, Kasten, and
  Dekel]{yatim2024space}
Danah Yatim, Rafail Fridman, Omer Bar-Tal, Yoni Kasten, and Tali Dekel.
\newblock Space-time diffusion features for zero-shot text-driven motion
  transfer.
\newblock In \emph{CVPR}, 2024.

\bibitem[Yin et~al.(2023)Yin, Wu, Liang, Shi, Li, Ming, and
  Duan]{yin2023dragnvwa}
Shengming Yin, Chenfei Wu, Jian Liang, Jie Shi, Houqiang Li, Gong Ming, and Nan
  Duan.
\newblock {DragNUWA: Fine-Grained Control in Video Generation by Integrating
  Text, Image, and Trajectory}.
\newblock \emph{arXiv preprint arXiv:2308.08089}, 2023.

\bibitem[Yuan et~al.(2024)Yuan, Liu, and Xu]{yuan2024bridging}
Shen Yuan, Haotian Liu, and Hongteng Xu.
\newblock Bridging the gap between low-rank and orthogonal adaptation via
  householder reflection adaptation.
\newblock \emph{NeurIPS}, 37:\penalty0 113484--113518, 2024.

\bibitem[Zhang et~al.(2025{\natexlab{a}})Zhang, Paiss, Zada, Karnad, Jacobs,
  Pritch, Mosseri, Shou, Wadhwa, and Ruiz]{zhang2024recapture}
David~Junhao Zhang, Roni Paiss, Shiran Zada, Nikhil Karnad, David~E. Jacobs,
  Yael Pritch, Inbar Mosseri, Mike~Zheng Shou, Neal Wadhwa, and Nataniel Ruiz.
\newblock {ReCapture}: Generative video camera controls for user-provided
  videos using masked video fine-tuning.
\newblock In \emph{CVPR}, 2025{\natexlab{a}}.

\bibitem[Zhang et~al.(2024)Zhang, Wei, Jiang, Zhang, Zuo, and
  Tian]{zhang2024controlvideo}
Yabo Zhang, Yuxiang Wei, Dongsheng Jiang, Xiaopeng Zhang, Wangmeng Zuo, and
  Qi~Tian.
\newblock {ControlVideo: Training-Free Controllable Text-to-Video Generation}.
\newblock In \emph{ICLR}, 2024.

\bibitem[Zhang et~al.(2025{\natexlab{b}})Zhang, Long, Qiu, Pan, Liu, Yao, and
  Mei]{zhang2025motionpro}
Zhongwei Zhang, Fuchen Long, Zhaofan Qiu, Yingwei Pan, Wu~Liu, Ting Yao, and
  Tao Mei.
\newblock Motionpro: A precise motion controller for image-to-video generation.
\newblock In \emph{CVPR}, pages 27957--27967, 2025{\natexlab{b}}.

\bibitem[Zheng et~al.(2025)Zheng, Li, Zhou, and Li]{zheng2025realcam}
Guangcong Zheng, Teng Li, Xianpan Zhou, and Xi~Li.
\newblock Realcam-vid: High-resolution video dataset with dynamic scenes and
  metric-scale camera movements.
\newblock \emph{arXiv preprint arXiv:2504.08212}, 2025.

\bibitem[Zheng et~al.(2024)Zheng, Peng, Yang, Shen, Li, Liu, Zhou, Li, and
  You]{zheng2024open}
Zangwei Zheng, Xiangyu Peng, Tianji Yang, Chenhui Shen, Shenggui Li, Hongxin
  Liu, Yukun Zhou, Tianyi Li, and Yang You.
\newblock Open-sora: Democratizing efficient video production for all.
\newblock \emph{arXiv preprint arXiv:2412.20404}, 2024.

\bibitem[Zhou et~al.(2018)Zhou, Tucker, Flynn, Fyffe, and
  Snavely]{zhou2018stereo}
Tinghui Zhou, Richard Tucker, John Flynn, Graham Fyffe, and Noah Snavely.
\newblock Stereo magnification: Learning view synthesis using multiplane
  images.
\newblock \emph{ACM TOG (Proc. SIGGRAPH)}, 37\penalty0 (4), 2018.

\end{thebibliography}

\clearpage
\appendix
\section{Other Related Work}
\paragraph{Image-to-Video Diffusion Models.} Diffusion-based video generation has progressed rapidly from early UNet-based architectures~\cite{ho2022vdm, blattmann2023align, blattmann2023stable, guo2024animatediff} that append temporal attention layers to pretrained image diffusion backbones, to large-scale Diffusion Transformer (DiT) models~\cite{peebles2023scalable} that jointly model spatial and temporal dependencies through full spatio-temporal attention~\cite{yang2024cogvideox, kong2024hunyuanvideo, wan2025wan, videoworldsimulators2024}. Image-to-video (I2V) models such as Stable Video Diffusion~\cite{blattmann2023stable} and CogVideoX~\cite{yang2024cogvideox} extend this capability by conditioning generation on a reference frame, enabling content-consistent animation from a single image. 
Despite their architectural differences, both UNet-based and DiT-based video diffusion models are initialized 
from Gaussian noise whose correlation across frames can be influential to the temporal consistency of the generated output~\cite{ge2023pyoco}. Our method exploits this property by modulating the noise with optical flow to induce structured temporal correlations before denoising begins, without modifying the denoising 
architecture.

\paragraph{Application of Householder Reflections in Deep Learning.}
Householder reflections provide a numerically stable and efficient parameterization of orthogonal operators, with a growing range of applications in deep learning. Early work uses them to enforce orthogonality in RNN transition matrices as a remedy for vanishing and exploding 
gradients~\cite{mhammedi2017efficient}. In generative modeling, products of Householder reflections serve as volume-preserving transformations in normalizing flows for VAEs~\cite{tomczak2016improving, hoogeboom2019emerging}. More recently, they have been adopted for parameter-efficient fine-tuning, where orthogonal weight updates preserve hyper-spherical energy during adaptation~\cite{qiu2023controlling, yuan2024bridging, arcas2025hoft, dong2024efficient}, and for positional encoding, where data-dependent Householder products replace static RoPE rotations to improve positional expressiveness~\cite{yang2025path, liu2025rethinking}. In all of these settings, Householder products operate on weight matrices, transition operators, or 
attention mechanisms. Our work applies them to the diffusion noise, where orthogonality guarantees that each frame's noise remains exactly i.i.d.\ Gaussian after noise modulation.

\section{Preliminaries}
\label{supp:prelim}

\paragraph{Latent Video Diffusion.}
We build on CogVideoX~\cite{yang2024cogvideox}, which encodes a video into a latent
$X_{0} \in \mathbb{R}^{\hat{N} \times C \times \hat{H} \times \hat{W}}$ with a 3D VAE and learns to
reverse the forward process $X_{t} = \alpha_{t} X_{0} + \sigma_{t}\,\boldsymbol{\epsilon}$,
$\boldsymbol{\epsilon} \sim \mathcal{N}(\mathbf{0}, \mathbf{I})$, where $\alpha_{t}^{2} + \sigma_{t}^{2} = 1$.
A DiT denoiser $D_{\theta}(X_{t}, t, c)$, conditioned on the first frame and the text prompt $c$, is
trained with the $v$-prediction objective
\begin{equation}
    \mathcal{L} = \mathbb{E}_{X_{0}, \boldsymbol{\epsilon}, t}\,\bigl\| D_{\theta}(X_{t}, t, c) - (\alpha_{t}\boldsymbol{\epsilon} - \sigma_{t} X_{0}) \bigr\|^{2},
\end{equation}
and sampling starts from $X_{T} = \boldsymbol{\epsilon}$. Noise-based motion-control methods, including noise warping and ours,
replace $\boldsymbol{\epsilon}$ by structured noise both in the training objective and as the initial
noise at inference, leaving the architecture and loss unchanged.

\section{Fine-tuning with Motion-Embedded Noise}
\label{supp:training}

\paragraph{Training Procedure.}
For each training video clip, we estimate pairwise forward optical flows with RAFT~\cite{teed2020raft}, construct $\tilde{Z}_{\gamma}$ via Algorithm~\ref{alg:noise}, and replace the standard i.i.d.\ 
initialization with $\tilde{Z}_{\gamma}$. The denoising objective and all other aspects 
of the training pipeline remain unchanged. At inference, the same noise 
construction is applied from a reference motion source, which can be obtained from a 
reference video, a predefined camera trajectory, or user-specified drag 
interactions on the initial frame.

\paragraph{Noise Degradation.}
Following GWTF~\cite{burgert2025go}, we interpolate between fully structured 
and fully unstructured noise via a degradation parameter $\gamma \in [0,1]$:
\begin{equation}
    \tilde{Z}_{\gamma}
    = \frac{(1-\gamma)\;\tilde{Z} + \gamma\;\boldsymbol{\zeta}}
           {\sqrt{(1-\gamma)^{2} + \gamma^{2}}},
    \qquad \boldsymbol{\zeta} \sim \mathcal{N}(\mathbf{0}, \mathbf{I}),
    \label{eq:degradation}
\end{equation}
where $\boldsymbol{\zeta}$ has the same shape as $\tilde{Z}$, $\gamma \to 0$ retains full motion conditioning and $\gamma \to 1$ 
approaches standard Gaussian noise. Since $\boldsymbol{\zeta}$ is drawn independently of $\tilde{Z}$ and the denominator normalizes the variance, every frame of $\tilde{Z}_{\gamma}$ remains exactly i.i.d.\ $\mathcal{N}(\mathbf{0}, \mathbf{I})$ across spatial locations and channels for any $\gamma$, so \cref{prop:gaussianity} carries over to the noise the model actually receives.
During training, $\gamma$ is sampled uniformly from $[0,1]$; at inference 
it is a user-controlled parameter governing the strength of motion adherence.

\section{Evaluation Protocol}
\label{supp:eval}
\paragraph{DAVIS (motion transfer).} We estimate forward optical flows on each reference video with RAFT~\cite{teed2020raft} and use them to construct $\tilde{Z}_{\gamma}$. Motion adherence is measured by comparing RAFT flows of the generated video against those of the reference video (Flow EPE), together with CoTracker~\cite{karaev2024cotracker} mIoU and PSNR.

\paragraph{MC-Bench (fine-grained and object-level control).} For each sample, we render a motion video from the provided mask and drag trajectory, estimate its flow with RAFT, and restrict the flow to the masked region.

\paragraph{DL3DV (camera control).} Following TTM~\cite{singer2025time}, we convert each camera trajectory into the corresponding optical flow. For ATE, camera poses are estimated with COLMAP~\cite{schoenberger2016sfm} on both the generated and the reference video, and the normalized error is computed between the two trajectories.

\paragraph{Baseline inputs.} All flow-conditioned methods (GWTF and ours) receive identical input flows, while the remaining baselines are conditioned on the same reference videos.

\section{Extended Details of Spatial Gaussianity Preservation}
\subsection{Proof of Proposition~\ref{prop:gaussianity}}
\label{supp:proof}

To prove \cref{prop:gaussianity}, we establish its two claims using the unrolled form 
$\tilde{\mathbf{z}}_{n}(\hat{x}) = \mathbf{P}_{n}(\hat{x})\,
\mathbf{z}_{0}(\hat{x})$, where 
$\mathbf{P}_{n}(\hat{x}) = \mathbf{Q}_{n}(\hat{x})\cdots\mathbf{Q}_{1}(\hat{x})$.

\begin{proof}\leavevmode

\paragraph{Marginal Gaussianity.}
Each $\mathbf{Q}_{j}(\hat{x}) = \prod_{k=1}^{K}\mathbf{R}(\mathbf{v}_{k}^{(j)}
(\hat{x}))$ is a product of Householder reflections, each of which is 
orthogonal by construction. Since a finite product of orthogonal matrices is orthogonal, 
$\mathbf{Q}_{j}(\hat{x})^{\top}\mathbf{Q}_{j}(\hat{x}) = \mathbf{I}_{C}$ 
for all $j$, and hence 
$\mathbf{P}_{n}(\hat{x})^{\top}\mathbf{P}_{n}(\hat{x}) = \mathbf{I}_{C}$.

For any orthogonal $\mathbf{P}$ and 
$\mathbf{z} \sim \mathcal{N}(\mathbf{0}, \mathbf{I}_{C})$:
\begin{equation}
    \mathbf{P}\mathbf{z} \sim \mathcal{N}(\mathbf{0}, 
    \mathbf{P}\mathbf{I}_{C}\mathbf{P}^{\top}) 
    = \mathcal{N}(\mathbf{0}, \mathbf{P}\mathbf{P}^{\top}) 
    = \mathcal{N}(\mathbf{0}, \mathbf{I}_{C}).
\end{equation}
Applying this to $\tilde{\mathbf{z}}_{n}(\hat{x}) = \mathbf{P}_{n}(\hat{x})\,
\mathbf{z}_{0}(\hat{x})$ yields 
$\tilde{\mathbf{z}}_{n}(\hat{x}) \sim \mathcal{N}(\mathbf{0}, \mathbf{I}_{C})$ 
for all $n$ and $\hat{x}$. For odd $K$, $\det\mathbf{Q}_{j}(\hat{x}) = -1$; this does not affect \cref{prop:gaussianity}, since any orthogonal matrix preserves $\mathcal{N}(\mathbf{0}, \mathbf{I}_{C})$, and the $K=1$ results in \cref{tab:rank_ablation} confirm that odd $K$ works in practice.

\paragraph{Spatial independence.}
The transformation at each location $\hat{x}$ acts solely on 
$\mathbf{z}_{0}(\hat{x})$ without coupling to neighboring locations. 
For any two distinct locations $\hat{x} \neq \hat{x}'$:
\begin{equation}
    \operatorname{Cov}\!\left[
        \tilde{\mathbf{z}}_{n}(\hat{x}),\,
        \tilde{\mathbf{z}}_{n}(\hat{x}')
    \right]
    = \mathbf{P}_{n}(\hat{x})\,
      \operatorname{Cov}\!\left[\mathbf{z}_{0}(\hat{x}),\,
      \mathbf{z}_{0}(\hat{x}')\right]
      \mathbf{P}_{n}(\hat{x}')^{\top}
    = \mathbf{P}_{n}(\hat{x})\,\mathbf{0}\,
      \mathbf{P}_{n}(\hat{x}')^{\top}
    = \mathbf{0},
\end{equation}
where $\operatorname{Cov}[\mathbf{z}_{0}(\hat{x}), 
\mathbf{z}_{0}(\hat{x}')] = \mathbf{0}$ follows from the spatial 
independence of the initial noise. Since the vectors 
$\tilde{\mathbf{z}}_{n}(\hat{x})$ are jointly Gaussian (each is a linear function of the initial noise) with zero cross-location 
covariance, they are mutually independent.
\end{proof}

\subsection{Cross-Frame Covariance Derivation}
\label{supp:covariance}

We derive the full covariance structure of the motion-embedded noise. For frames $n$, $m$ and latent locations $\hat{x}$, $\hat{x}'$:
\begin{align}
    \operatorname{Cov}\!\left[
        \tilde{\mathbf{z}}_{n}(\hat{x}),\,
        \tilde{\mathbf{z}}_{m}(\hat{x}')
    \right]
    &= \mathbb{E}\!\left[
        \mathbf{P}_{n}(\hat{x})\,\mathbf{z}_{0}(\hat{x})\,
        \mathbf{z}_{0}(\hat{x}')^{\top}\,\mathbf{P}_{m}(\hat{x}')^{\top}
    \right] \nonumber \\
    &= \mathbf{P}_{n}(\hat{x})\,
       \mathbb{E}\!\left[
           \mathbf{z}_{0}(\hat{x})\,\mathbf{z}_{0}(\hat{x}')^{\top}
       \right]
       \mathbf{P}_{m}(\hat{x}')^{\top}
    = \delta_{\hat{x}\hat{x}'}\,\mathbf{P}_{n}(\hat{x})\,\mathbf{P}_{m}(\hat{x}')^{\top},
    \label{eq:full_cov}
\end{align}
where $\delta_{\hat{x}\hat{x}'}$ is the Kronecker delta, following from the zero mean and spatial independence of $\mathbf{z}_{0}$. Noise at distinct locations is therefore uncorrelated across all frames, and at the same location the covariance reduces to \cref{eq:cross_frame}; for $n > m$,
\begin{equation}
    \operatorname{Cov}\!\left[
        \tilde{\mathbf{z}}_{n}(\hat{x}),\,
        \tilde{\mathbf{z}}_{m}(\hat{x})
    \right]
    = \mathbf{P}_{n}(\hat{x})\,\mathbf{P}_{m}(\hat{x})^{\top}
    = \mathbf{Q}_{n}(\hat{x})\cdots\mathbf{Q}_{m+1}(\hat{x}),
    \label{eq:cross_frame_supp}
\end{equation}
an orthogonal matrix determined only by the flows between frames $m$ and $n$ at $\hat{x}$. Its Frobenius norm is $\sqrt{C}$ for all $n$, $m$ and $\hat{x}$, so the strength of the cross-frame dependence is reflected not in its norm but in its alignment with the identity: the average channel-wise correlation $\frac{1}{C}\operatorname{tr}\bigl(\mathbf{P}_{n}(\hat{x})\mathbf{P}_{m}(\hat{x})^{\top}\bigr) = \frac{1}{C}\mathbb{E}\bigl[\tilde{\mathbf{z}}_{n}(\hat{x})^{\top}\tilde{\mathbf{z}}_{m}(\hat{x})\bigr] \in [-1, 1]$ equals $1$ if and only if $\mathbf{Q}_{n}(\hat{x})\cdots\mathbf{Q}_{m+1}(\hat{x}) = \mathbf{I}_{C}$.

\subsection{Experimental Results}

 \begin{table}[t]   
  \centering   
  \caption{Comparison of Gaussianity preservation and computational efficiency between methods that produce temporally correlated noise (1024$\times$1024 resolution, $C=16$) driven by optical flow; statistics for our method are computed on $\tilde{Z}$ (i.e., $\gamma = 0$), and all GPU times are measured on an NVIDIA RTX PRO 6000 Blackwell GPU.}                                                                                                                                                                      
  \label{tab:gaussianity}
  \resizebox{\textwidth}{!}{%
  \begin{tabular}{lcccccccc}                                                                                                                                                                                                                               
  \toprule                                                                                                                                                                                                                    \multirow{2}{*}{Method} 
     
  & Moran's I & Moran's I & K-S Test & K-S Test  & Std. & Skewness & Ex. Kurtosis  & GPU Time \\                                                             
  & ($\approx 0$) & p-value$\uparrow$ & (index) $\downarrow$ & p-value $\uparrow$ & ($\approx 1.0$) & ($\approx 0$) & ($\approx 0$) & ms/Latent$\downarrow$ \\                 
  \midrule                                                                                                                                                                                                                                                
  \multicolumn{9}{l}{\textit{Warping + Interpolation Operation}} \\
  
  Nearest         & 0.1150 & 0.012 & 0.0107 & 0.115 & 1.0015 & 0.0016 & -0.0116 & - \\                                                                                                                                                      
  Bilinear        & 0.6218 & 0.008 & 0.3318 & 0.000 & 0.2179 & 0.0720 & 0.4753   & - \\                                                                                                                                                    
  Bicubic         & -0.0671 & 0.020 & 0.0670 & 0.000 & 1.0489 & 0.0075 & 0.2194   & - \\
      
  PYoCo~\cite{ge2023pyoco}  & 0.0002 & 0.478 &0.0033 & 0.512 & 1.000 & 0.0005 & 0.0000 & - \\
  \midrule                                                                                                                                                         
  \multicolumn{9}{l}{\textit{Optical Flow Driven Noise Warping Methods}} \\    
  CaV~\cite{zhang2024controlvideo2} & 0.6170 & 0.004 & 0.3379 & 0.000 & 0.2060 & 0.0590 & 4.065 & - \\

  HIWYN~\cite{chang2024warped}$_{L=3}$ & 0.0001 & 0.502 & 0.0037 & 0.422 & 0.9957 & 0.0007 & -0.0001 & 38.00\\ 
  
  GWTF~\cite{burgert2025go}   & 0.0002 & 0.507 & 0.0035 & 0.494 & 0.9997 & -0.0006 & 0.0037 & 7.63 \\                                                                                                                                                          
  InfRes~\cite{deng2025infiniteresolution} & 0.0003 & 0.512 & 0.0034 & 0.496 & 1.0001 & 0.0004 & 0.0011 & 23.94 \\                                                                                                                                            
  \midrule                                                                                                                                                                                      
  \multicolumn{9}{l}{\textit{Ours (HME)}} \\                                             
  $K=2$  & 0.0003 & 0.482 & 0.0032 & 0.535 & 0.9998 & -0.0006 & 0.0014 & \textbf{2.80} \\                                    
  $K=4$  & 0.0003 & 0.475 & 0.0032 & 0.531 & 0.9999  & -0.0005 & 0.0014 & \underline{5.18} \\                                                         
  $K=6$  & 0.0003 & 0.466 & 0.0033 & 0.527 & 1.0001 & -0.0007 & 0.0015 & 7.59 \\                                                     
  $K=8$  & 0.0003 & 0.493 & 0.0033 & 0.528 & 0.9998 & -0.0004 & 0.0011 & 10.01 \\                                                 
  $K=12$ & 0.0002 & 0.534 & 0.0032 & 0.546 & 1.0001 & -0.0005 & 0.0013 & 14.78 \\                                                                                                                                            
  \bottomrule                                                                                                                                                                                                                                                
  \end{tabular}%
  }        
\end{table}       

\paragraph{Metrics and Baselines.}
 To evaluate the Gaussianity preservation ability of our method, we follow InfRes~\cite{deng2025infiniteresolution} and GWTF~\cite{burgert2025go} to present Moran's I and Kolmogorov-Smirnov (K-S) test results. Moran's I measures spatial autocorrelation of warped noise 
 and the K-S test assesses the distributional fidelity of the warped noise against a standard normal distribution.
 We further report noise Std., skewness 
 and excess kurtosis 
 to characterize higher-order distributional properties such as the symmetry and shape of the resulting distribution compared to normal distribution. We also evaluate the noise processing efficiency of all methods using GPU time. Baselines mainly include optical flow driven warping methods (CaV, HIWYN~\cite{chang2024warped}, GWTF~\cite{burgert2025go}, InfRes~\cite{deng2025infiniteresolution}) designed for Gaussianity preservation. These benchmarks operate on the noise tensor alone and run at $1024 \times 1024$, $C = 16$, on an NVIDIA RTX PRO 6000 Blackwell GPU; this setting only measures noise properties, computational overhead and memory, and is independent of the $480 \times 720$ video generation experiments in \cref{sec:experiments}.

\paragraph{Performance Analysis.}
As shown in~\cref{tab:gaussianity}, \textbf{HME} is successful in generating spatially Gaussian noise with Moran's I index values close to 0, with corresponding p-values ranging from 0.466 to 0.534, indicating no significant spatial autocorrelation, an important property of i.i.d. noise. Small K-S index values and p-values above 0.05 indicate the spatial distribution of resulting noise matches a standard normal distribution. \textbf{HME} obtains a K-S statistic of $\sim0.0032$ and p-value of $\sim0.53$ on average, demonstrating the distributional fit to a standard normal distribution. Furthermore, our method achieves near 1 Std., negligible Skewness and Excess Kurtosis, suggesting substantial normality. In contrast, all interpolation-based operations introduce significant spatial autocorrelation (Moran's I p-values of $0.008$--$0.020$); bilinear and bicubic interpolation additionally fail the K-S test, while nearest-neighbor sampling preserves the marginal distribution (K-S p-value $= 0.115$) but not the spatial independence of the noise. 

Meanwhile, dedicated flow-based methods also recover near-zero Moran's I, acceptable K-S p-values, and negligible skewness and excess kurtosis, but at runtimes of $7.63$--$38.00$ ms/latent. Our method reduces the GPU runtime to $\mathbf{2.80}$ ms/latent at $K=2$, a $2.7\times$ speedup over GWTF and $13.6\times$ over HIWYN. All our variants ($K=2$ to $12$) satisfy the Gaussianity criteria with runtime scaling gracefully from $2.80$ to $14.78$ ms/latent. 

\section{Extra Qualitative Comparisons}
\label{supp:extra}
Video versions of the comparisons in this paper, along with additional results, are provided in the Supplementary Material; differences between methods are easier to spot in the videos.
\subsection{Motion Transfer}
\begin{figure}[h!]
   \centering
   \includegraphics[width=\linewidth]{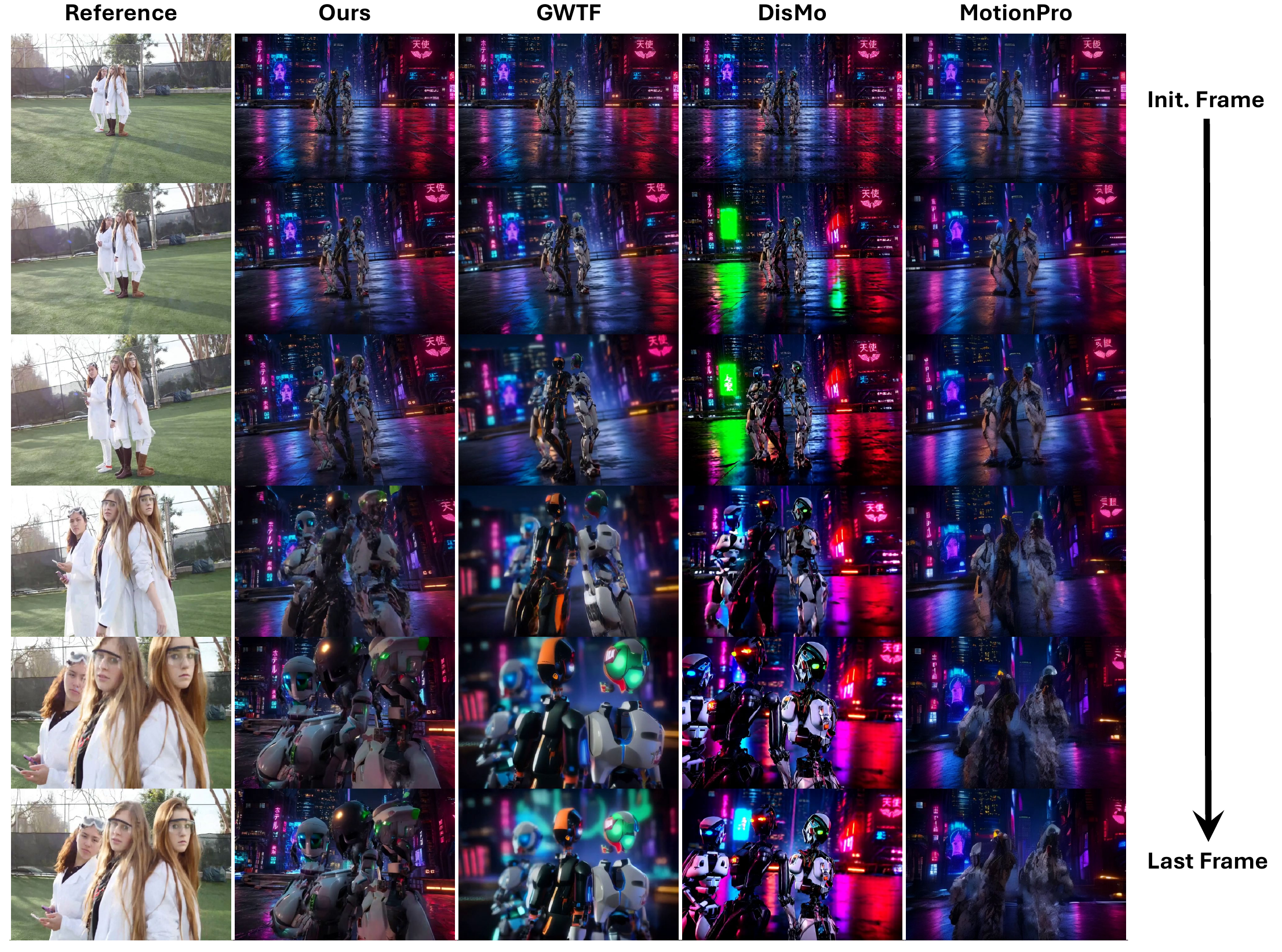}
   \caption{Motion transfer example: we demonstrate strong motion adherence to the reference video while maintaining high generation quality and fidelity. \textit{Prompt used: "Cyberpunk rooftop at night, wet reflective ground, neon city skyline, three humanoid robots standing slightly left of center in a triangular group pose, aligned like three girls in an open field. White-blue robot, black-orange robot, white-red-green robot. Wide composition, large empty foreground and right-side space, glowing Japanese signs, holographic billboards, photorealistic, cinematic, ultra detailed."}}
  \label{fig:mt1}
\end{figure}

For clarity, we show qualitative motion-transfer comparisons with GWTF, DisMo and MotionPro in~\cref{fig:mt1}; quantitative comparisons with all baselines, including TTM, are given in~\cref{tab:davis}.

\subsection{Qualitative Comparisons From MC-Bench}
Following the experiment described in \cref{sec:gen}, we present additional results beyond those shown in
\cref{fig:Qualitative1}, further illustrating our method’s performance against leading approaches on the MC-Bench
dataset.

\begin{figure}[h!]
   \centering
   \includegraphics[width=\linewidth]{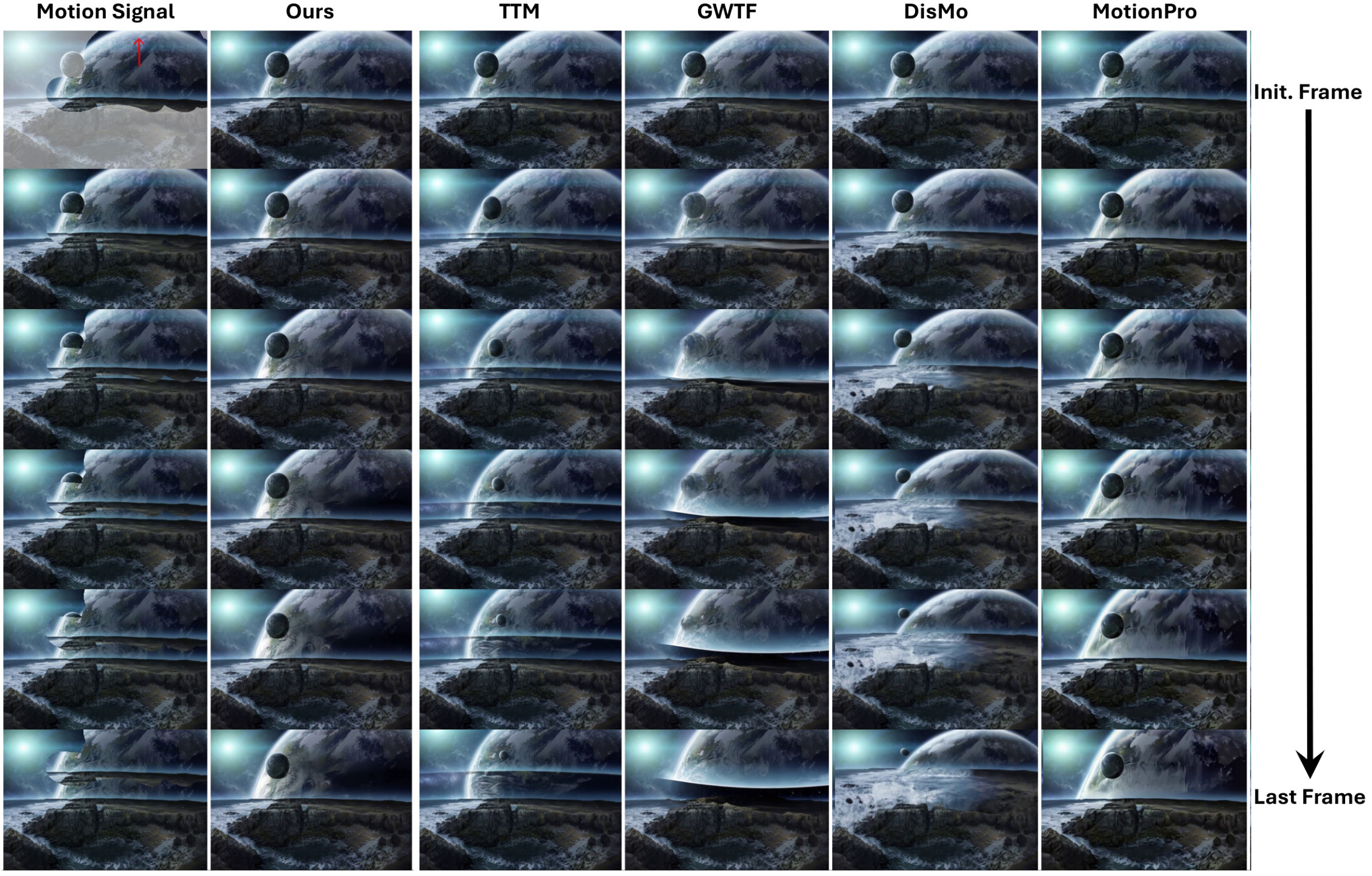}
   \caption{Example from MC-Bench: we highlight the disentanglement between the moving subject (masked planet) and the background in our method.}
\end{figure}

\begin{figure}[h!]
   \centering
   \includegraphics[width=\linewidth]{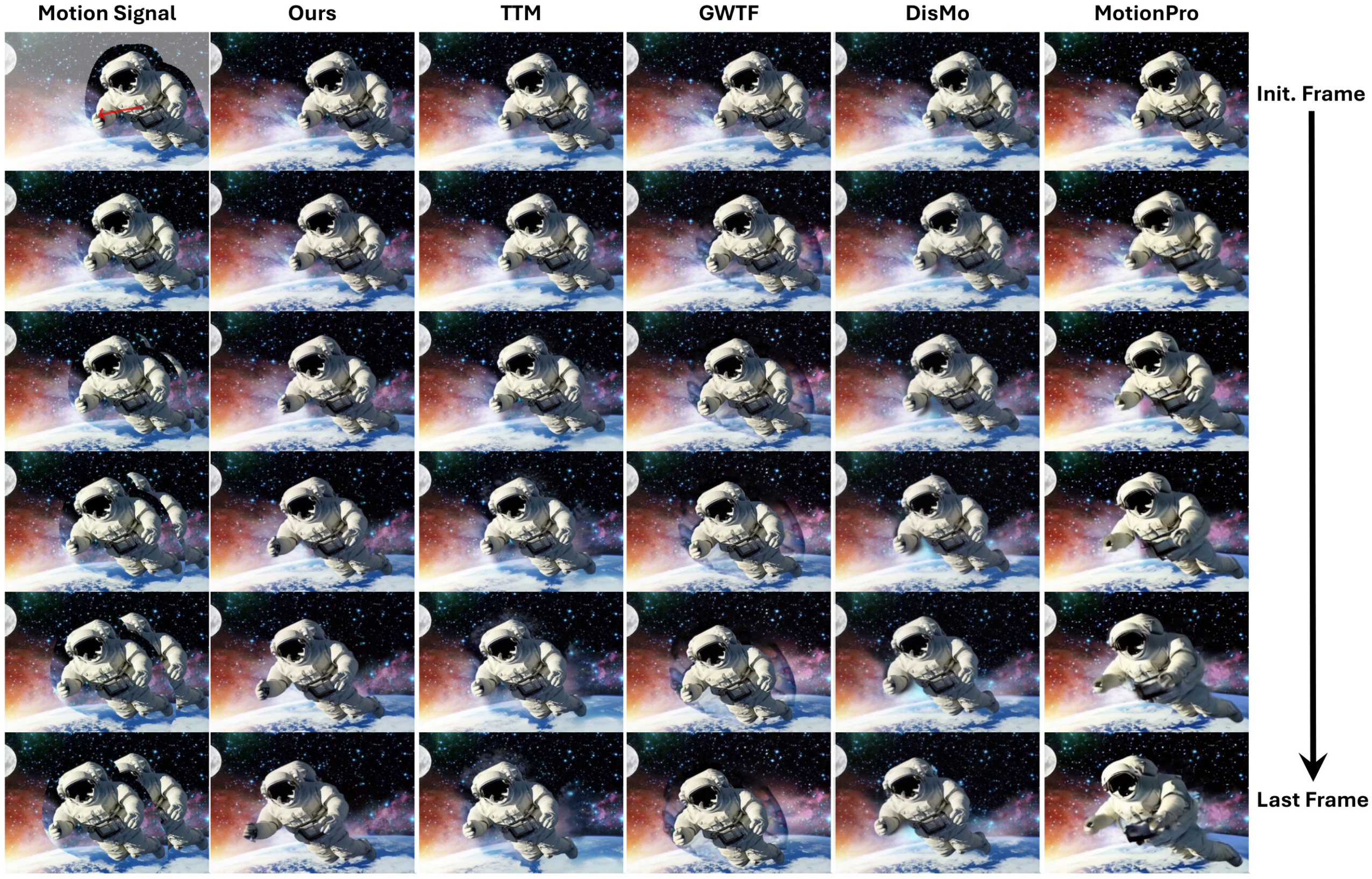}
   \caption{Example from MC-Bench: we highlight the disentanglement between the moving astronaut and the background in our method.}
\end{figure}

\clearpage
\subsection{Qualitative Comparisons From DAVIS}
Following the experiment described in \cref{sec:gen}, we present additional results beyond those shown in \cref{fig:Qualitative1}, further illustrating our method’s performance against leading approaches on the DAVIS dataset.

\begin{figure}[h!]
   \centering
   \includegraphics[width=\linewidth]{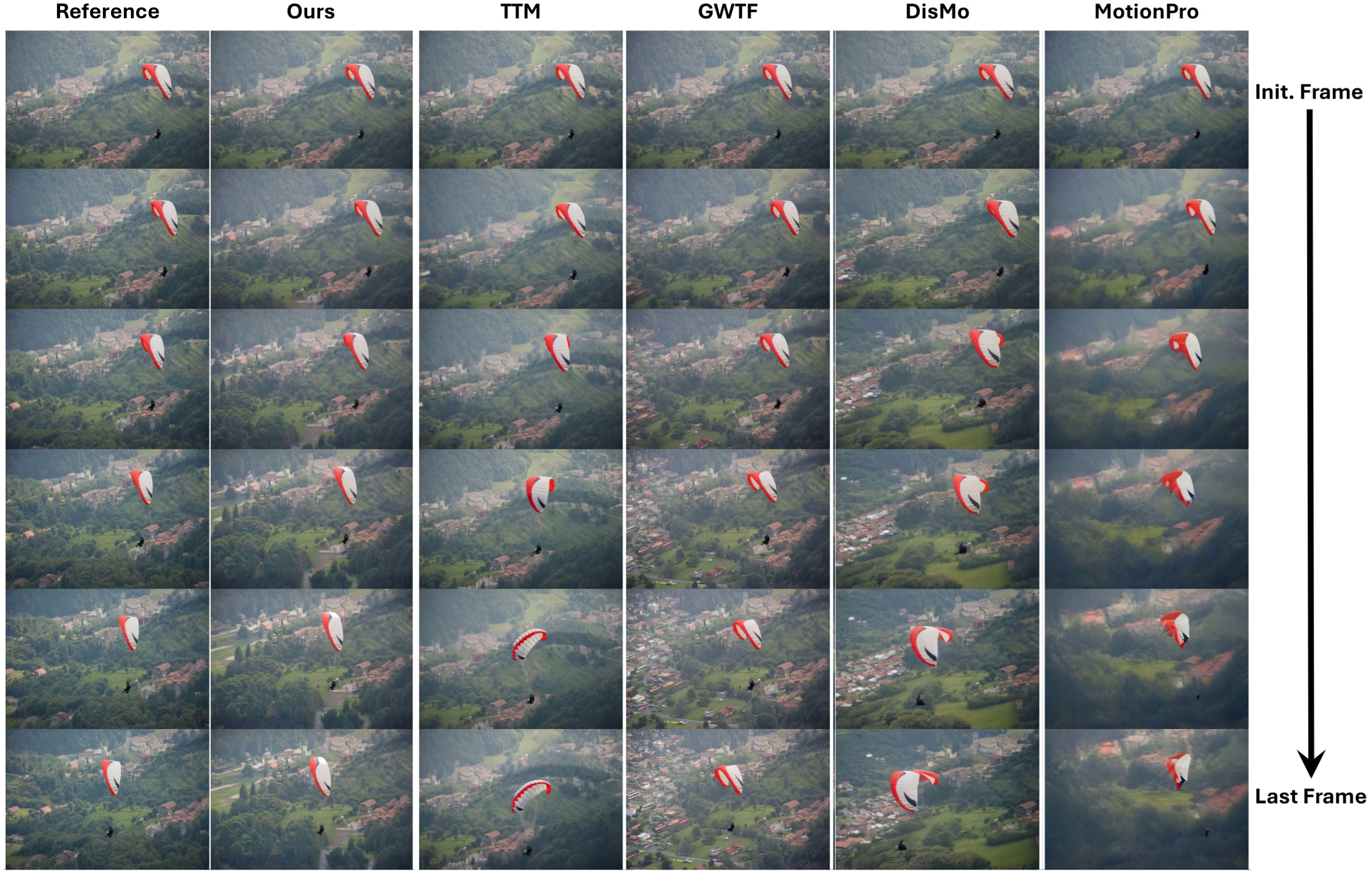}
   \caption{Example from DAVIS: our method shows better motion adherence.}
\end{figure}

\begin{figure}[h!]
   \centering
   \includegraphics[width=\linewidth]{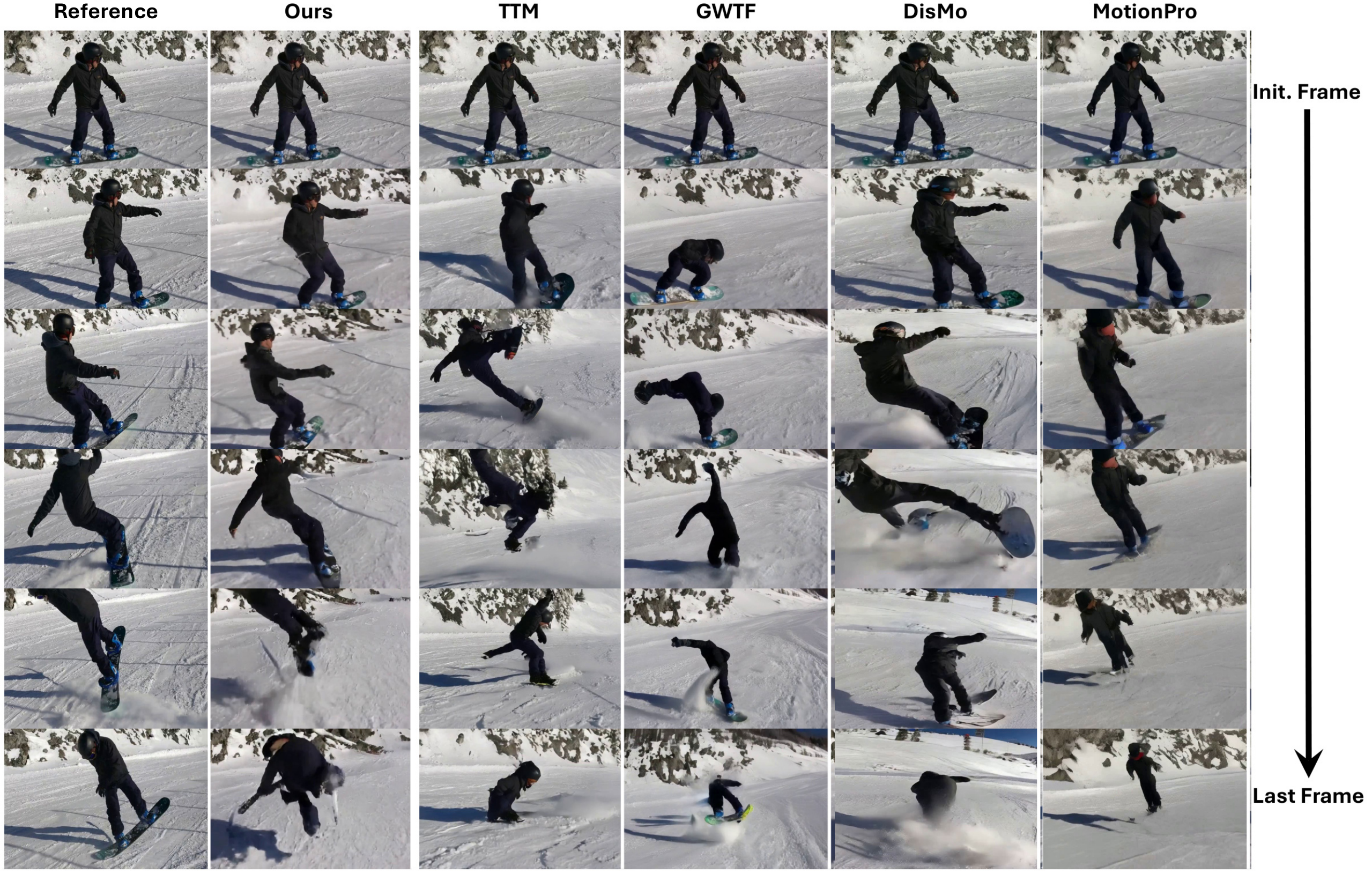}
   \caption{Example from DAVIS: our method shows better motion adherence.}
\end{figure}

\clearpage
\subsection{Qualitative Comparisons From DL3DV}
Following the experiment described in \cref{sec:gen}, we present additional camera-control results beyond those shown in \cref{fig:camcontrol}.

\begin{figure}[h!]
   \centering
   \includegraphics[width=\linewidth]{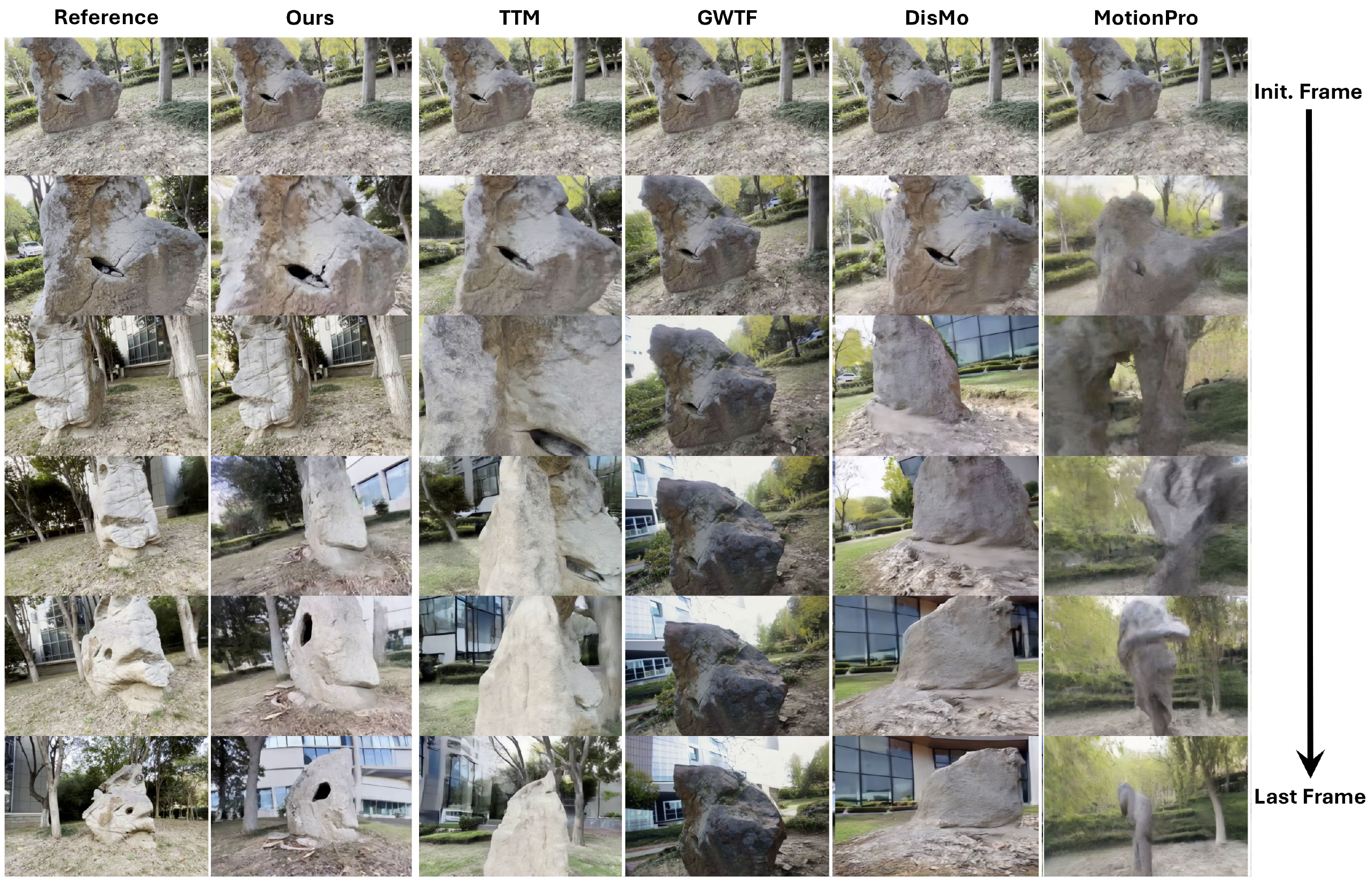}
   \caption{Example from DL3DV: our method shows better camera motion adherence under optical flow control.}
\end{figure}

\begin{figure}[h!]
   \centering
   \includegraphics[width=\linewidth]{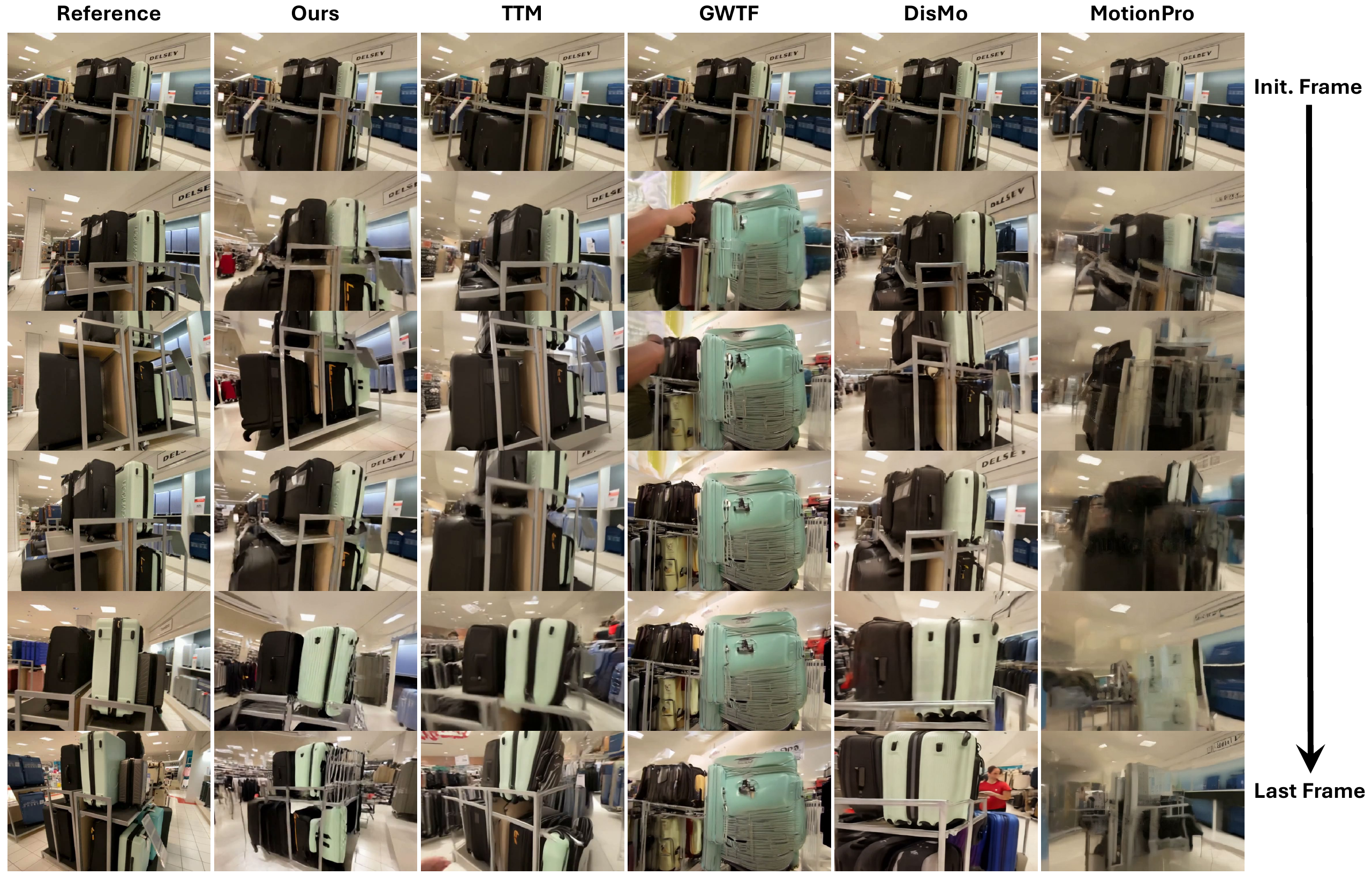}
   \caption{Example from DL3DV: our method shows better camera motion adherence under optical flow control.}
\end{figure}

\clearpage
\section{Qualitative Results for Different Noise Degradation Levels}
\label{supp:ablat}
We qualitatively ablate the noise degradation level $\gamma$, as analyzed in \cref{sec:ablat}.
\begin{figure}[h!]
   \centering
   \includegraphics[width=\linewidth]{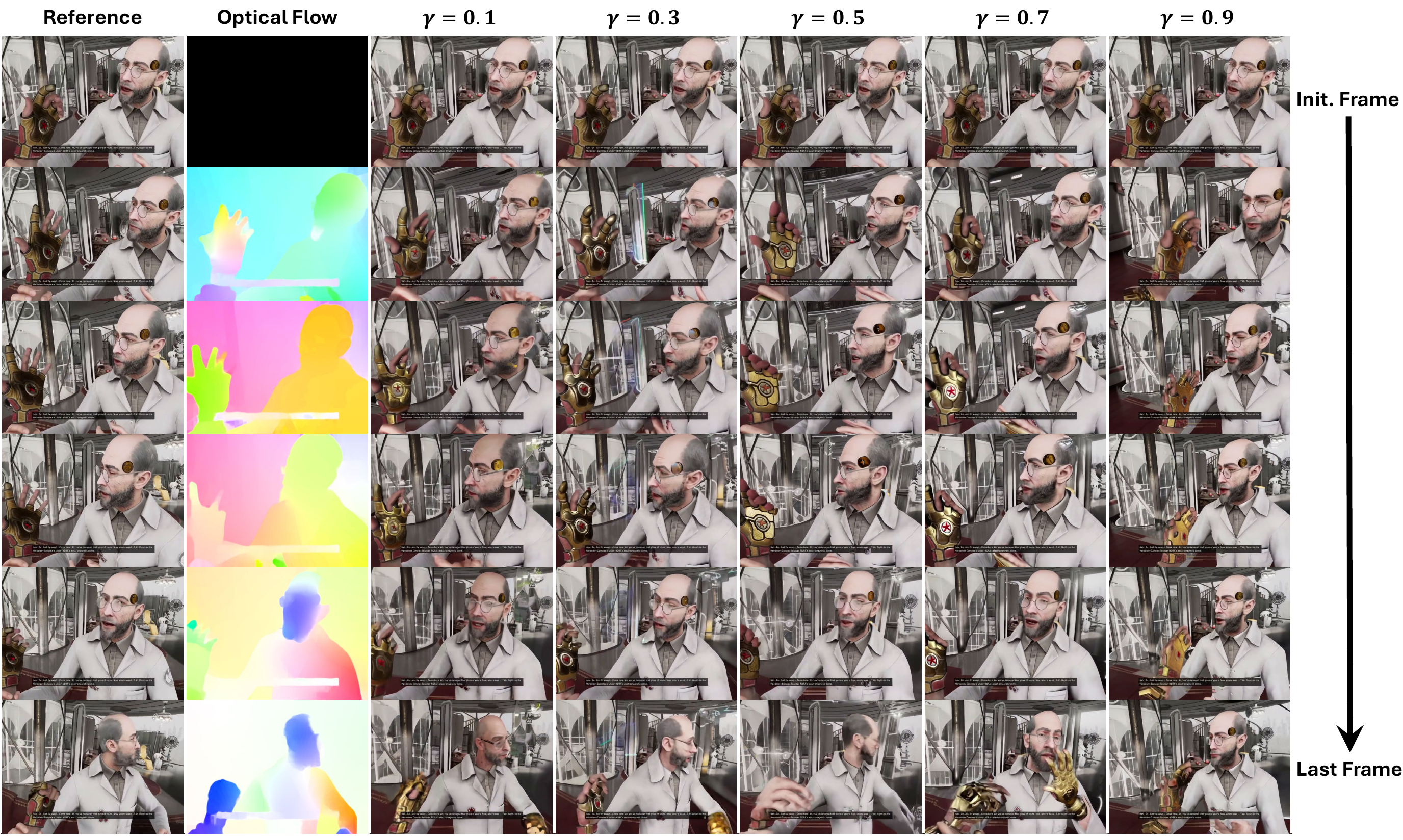}
   \caption{The effect of noise degradation level $\gamma$ on generated videos. As degradation decreases, the video adheres more to the input flow.}
\end{figure}

\end{document}